\documentclass[runningheads]{llncs}
\usepackage{graphicx}

\usepackage{tikz}
\usepackage{comment} 
\usepackage{amsmath,amssymb,bm} 
\usepackage{color}
\usepackage{booktabs}
\usepackage{multirow}
\usepackage{subfig}
\usepackage{makecell}
\usepackage[pagebackref=true,breaklinks,letterpaper=true,colorlinks=true,bookmarks=false,]{hyperref}

\begin{document}
\def\eg{e.g.} \def\Eg{E.g.}
\def\ie{i.e.} \def\Ie{I.e.}
\def\cf{c.f.} \def\Cf{C.f.}
\def\etc{etc.} \def\vs{vs.}
\def\wrt{w.r.t.} \def\dof{d.o.f.}
\def\etal{et al.}

\pagestyle{headings}
\mainmatter
\def\ECCVSubNumber{1834}  

\title{Sketching Image Gist: Human-Mimetic Hierarchical Scene Graph Generation} 

\titlerunning{Sketching Image Gist}
%
\author{Wenbin Wang\inst{1,2}\orcidID{0000-0002-4394-0145} \and
Ruiping Wang\inst{1,2}\orcidID{0000-0003-1830-2595} \and
Shiguang Shan\inst{1,2}\orcidID{0000-0002-8348-392X} \and
Xilin Chen\inst{1,2}\orcidID{0000-0003-3024-4404}}
\authorrunning{W. Wang et al.}
%
\institute{Key Laboratory of Intelligent Information Processing of Chinese Academy of Sciences (CAS), Institute of Computing Technology, CAS, Beijing, 100190, China \and
University of Chinese Academy of Sciences, Beijing, 100049, China
\email{wenbin.wang@vipl.ict.ac.cn}, 
\email{\{wangruiping, sgshan, xlchen\}@ict.ac.cn}}
\maketitle

\begin{abstract}
Scene graph aims to faithfully reveal humans' perception of image content. When humans analyze a scene, they usually prefer to describe image gist first, namely major objects and key relations in a scene graph. This humans' inherent perceptive habit implies that there exists a hierarchical structure about humans' preference during the scene parsing procedure. 
Therefore, we argue that a desirable scene graph should be also hierarchically constructed, and introduce a new scheme for modeling scene graph. Concretely, a scene is represented by a human-mimetic \textbf{H}ierarchical \textbf{E}ntity \textbf{T}ree (HET) consisting of a series of image regions. To generate a scene graph based on HET, we parse HET with a Hybrid Long Short-Term Memory (Hybrid-LSTM) which specifically encodes hierarchy and siblings context to capture the structured information embedded in HET. To further prioritize key relations in the scene graph, we devise a Relation Ranking Module (RRM) to dynamically adjust their rankings by learning to capture humans' subjective perceptive habits from objective entity saliency and size. Experiments  indicate that our method not only achieves state-of-the-art performances for scene graph generation, but also is expert in mining image-specific relations which play a great role in serving downstream tasks. 
\end{abstract}

\section{Introduction}

\begin{figure}[t]
\setlength{\abovecaptionskip}{-0.2cm}
\setlength{\belowcaptionskip}{-0.6cm}
\begin{center}
\includegraphics[width=1.0\linewidth]{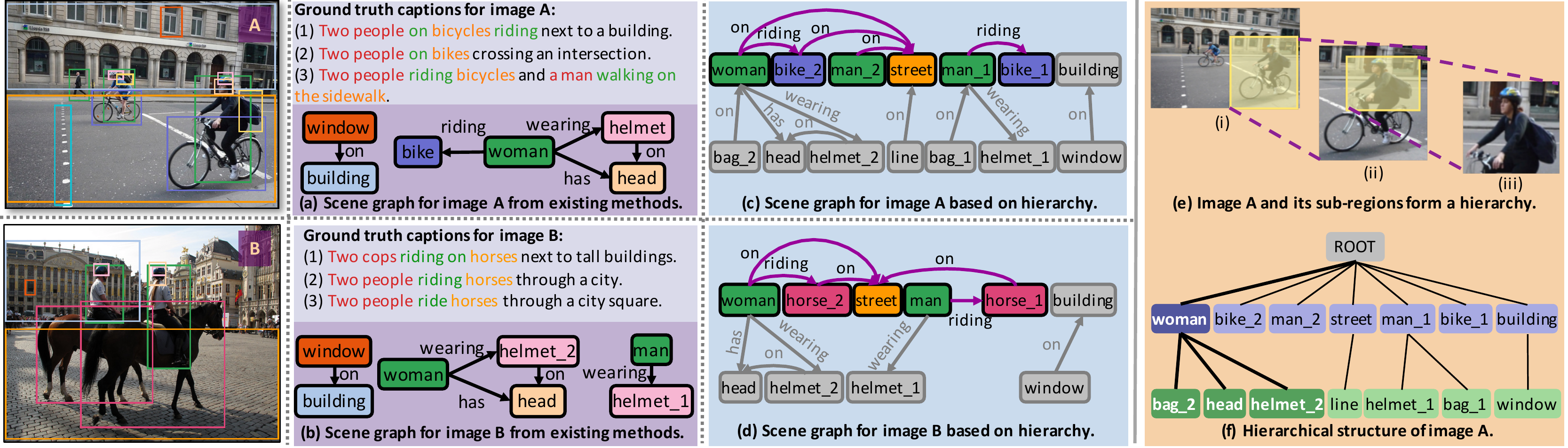}
\end{center}
\caption{Scene graphs from existing methods shown in (a) and (b) fail in sketching the image gist. The hierarchical structure about humans' perception preference is shown in (f), where the bottom left highlighted branch stands for the hierarchy in (e). The scene graphs in (c) and (d) based on hierarchical structure better capture the gist. Relations in (a) and (b), and purple arrows in (c) and (d), are top-5 relations, while gray ones in (c) and (d) are secondary.}
\label{fig:fig_intro}
\end{figure}

In an effort to thoroughly understand a scene, scene graph generation (SGG) \cite{johnson2015image,xu2017scene} in which objects and pairwise relations should be detected, has been on the way to bridge the gap between low-level recognition and high-level cognition, and contributes to tasks like image captioning \cite{wu2018image,lu2018neural,yao2018exploring}, VQA \cite{antol2015vqa,Tang_2019_CVPR}, and visual reasoning \cite{Shi_2019_CVPR}. While previous works \cite{xu2017scene,li2017scene,yang2018graph,li2018factorizable,zhang2017visual,Qi_2019_CVPR,Tang_2019_CVPR,Wang_2019_CVPR,zellers2018neural,Zhang2019graphical} have pushed this area forward, the generated scene graph may be still far from perfect, \eg, they seldom consider whether the detected relations are what humans want to convey from the image or not. As a symbolic representation of an image, the scene graph is expected to record the image content as complete as possible. More importantly, a scene graph is not just for being admired, but for supporting downstream tasks, such as image captioning, where a description is supposed to depict the major event in the image, or the namely \textbf{image gist}. This characteristic is also one of the humans' inherent habits when they parse a scene. Therefore, an urgently needed feature of SGG is to assess the relation importance and prioritize the relations which form the major events that humans intend to preferentially convey, \ie, \textbf{key relations}. This is seldom considered by existing methods. What's worse, the universal phenomenon of unbalanced distribution of relationship triplets in mainstream datasets exacerbates the problem that the major event cannot be found out. Let's study the quality of top relations predicted by existing state-of-the-art methods (\eg, \cite{zellers2018neural}) and check whether they are \lq\lq key\rq\rq\,\,or not.
In Figure \ref{fig:fig_intro}(a)(b), two scene graphs shown with top-5 relations for image A and B are mostly the same although major events in A and B are quite different. In other words, existing methods are deficient in mining image-specific relations, but biased towards trivial or self-evident ones (\eg, $\langle$\textit{woman}, \textit{has}, \textit{head}$\rangle$ can be obtained from commonsense without observing the image), which fail in conveying image gist (colored parts in ground truth captions in Figure \ref{fig:fig_intro}), and barely contribute to downstream tasks. 

Any pair of objects in a scene can be considered relevant, at least in terms of their spatial configurations. Faced with such a massive amount of relations, how do humans choose relations to  describe the images? Given picture (ii) in Figure \ref{fig:fig_intro}(e), a zoom-in sub-region of picture (i), humans will describe it with $\langle$\textit{woman}, \textit{riding}, \textit{bike}$\rangle$, since \textit{woman} and \textit{bike} belong to the same perceptive level and their interaction forms the major event in (ii). When it comes to picture (iii), the answers would be $\langle$\textit{woman}, \textit{wearing}, \textit{helmet}$\rangle$ and $\langle$\textit{bag}, \textit{on}, \textit{woman}$\rangle$, where \textit{helmet} and \textit{bag} are finer details of \textit{woman} and belong to an inferior perceptive level. It suggests that there naturally exists a hierarchical structure about humans' perception preference, as shown in Figure \ref{fig:fig_intro}(f).

Inspired by observations above, we argue that a desirable scene graph should be hierarchically constructed. Specifically, we represent the image with a human-mimetic Hierarchical Entity Tree (HET) where each node is a detected object and each one can be decomposed into a set of finer objects attached to it. To generate the scene graph based on HET, we devise Hybrid Long Short-Term Memory (Hybrid-LSTM) to encode both hierarchy and siblings context \cite{zellers2018neural,Tang_2019_CVPR} and capture the structured information embedded in HET, considering that important related pairs are more likely to be seen either inside a certain perceptive level or between two adjacent perceptive levels. We further intend to evaluate the performances of different models on key relation prediction but the annotations of key relations are not directly available from existing datasets. Therefore, we extend Visual Genome (VG) \cite{krishna2017visual} to VG-KR dataset which contains indicative annotations of key relations by drawing support from caption annotations in MSCOCO \cite{lin2014microsoft}. We devise a Relation Ranking Module to adjust the rankings of relations. It captures humans' subjective perceptive habits from objective entity saliency and size, and achieves ultimate performances on mining key relations.\footnote[1]{Source code and dataset are available at \href{http://vipl.ict.ac.cn/resources/codes}{\textcolor{blue}{http://vipl.ict.ac.cn/resources/codes}} or \href{https://github.com/Kenneth-Wong/het-eccv20.git}{\textcolor{blue}{https://github.com/Kenneth$-$Wong/het$-$eccv20.git}}. }

\section{Related Works}

\noindent{
\textbf{Scene graph generation (SGG) and Visual Relationship Detection (VRD)}, are the two most common tasks aiming at extracting interaction between two objects. In the field of VRD, various studies \cite{lu2016visual,dai2017detecting,li2017vip,zhang2017visual,yu2017visual,peyre2017weakly,zhang2017ppr,yin2018zoom,zhang2019large} mainly focus on detecting each relation triplet independently rather than describe the structure of the scene. The concept of scene graph is firstly proposed in \cite{johnson2015image} for image retrieval. Xu \etal\,\,\cite{xu2017scene} define SGG task and creatively devise message passing mechanism for scene graph inference. A series of succeeding works struggle to design various approaches to improve the graph representation. Li \etal\,\,\cite{li2017scene} induce image captions and object information to jointly address multitasks. \cite{zellers2018neural,Tang_2019_CVPR,Wang_2019_CVPR,Lin_2020_CVPR} draw support from useful context construction. Yang \etal\,\,\cite{yang2018graph} propose Graph-RCNN to embed the structured information. Qi \etal\,\,\cite{Qi_2019_CVPR} employ a self-attention module to embed a weighted graph representation. Zhang \etal\,\,\cite{Zhang2019graphical} propose contrastive losses to resolve the related pair configuration ambiguity. Zareian \etal\,\,\cite{Zareian_2020_CVPR} creatvely treat the SGG as an edge role assignment problem. Recently, some methods try to borrow advantages from using knowledge \cite{Chen_2019_CVPR,Gu_2019_CVPR} or causal effect \cite{Tang_2020_CVPR} to diversify the predicted relations. Liang \etal\,\,\cite{liang2019vrr} prune the dominant and easy-to-predict relations in VG to alleviate the annihilation problem of rare but meaningful relations.

\noindent{\textbf{Structured Scene Parsing}}, has been paid much attention in pursuit of higher-level scene understanding. \cite{socher2011parsing,sharma2015deep,lin2016deep,han2008bottom,zhu2011recursive,yao2019hierarchy} construct various hierarchical structures for their specific tasks. Unlike existing SGG studies that indiscriminately detect relations no matter whether they are concerned by humans or not, our work introduces the idea of hierarchical structure into SGG task, and try to give priority to detect key relations, then the trivial ones for completeness. 

\noindent{\textbf{Saliency \vs\,\,Image Gist.}} 
An extremely rich set of studies \cite{li2015visual,wang2015deep,liu2018picanet,wang2017stagewise,HouCvpr2017Dss,Zhang_2019_CVPR} focus on analyzing where humans gaze and find visually salient objects (high contrast of luminance, hue, and saturation, center position \cite{itti1998model,klein2011center,xie2012bayesian}, \etc). It's notable that the visually salient objects are related but not equal to objects involved in image gist. He \etal\,\,\cite{He_2019_ICCV} explore gaze data and find that only 48\% of fixated objects are referred in humans' descriptions about the image, while 95\% of objects referred in descriptions are fixated. It suggests that objects referred in a description (\ie, objects that humans think important and should form the major events / image gist) are almost visually salient and  reveal where humans gaze, but what humans fixate (\ie, visually salient objects) are not always what they want to convey. We provide some examples in the Appendix to help to understand this finding. Naturally, we need to emphasize that the levels in our HET reflect the perception priority level rather than the object saliency. 
Besides, this finding supports us to obtain the indicative annotations of key relations with the help of image caption annotations. 

\section{Proposed Approach}
\subsection{Overview}
The scene graph $\mathcal{G}=\{\mathcal{O}, \mathcal{R}\}$ of an image $\mathcal{I}$ contains a set of entities $\mathcal{O}=\{o_i\}_{i=1}^N$ and their pairwise relations $\mathcal{R}=\{r_k\}_{k=1}^M$. Each $r_k$ is a triplet $\langle o_i, p_{ij}, o_j\rangle$ where $p_{ij} \in \mathcal{P}$ and $\mathcal{P}$ is the set of all predicates.  As illustrated in Figure \ref{fig:framework}, our approach can be summarized into four steps. (i) We apply Faster R-CNN \cite{ren2015faster} with VGG16 \cite{simonyan2014very} backbone to detect all the entity proposals and each of them possesses its bounding box $\bm{b}_i\in \mathbb{R}^4$, 4,096-dimensional visual feature $\bm{v}_i$, and the class probability vector $\bm{q}_i$ from the softmax output.  (ii) In Section \ref{sec:tree}, HET is constructed by organizing the detected entities according to their perceptive levels. (iii) In Section \ref{sec:hlstm}, we design the Hybrid-LSTM network to parse HET, which firstly encodes the structured context then decodes it for graph inference. (iv) In Section \ref{sec:RRM}, we improve the scene graph generated in (iii) with our devised RRM which further adjusts the rankings of relations and shifts the graph focus to the relations between entities that are close to top perceptive levels of HET.

\begin{figure}[t]
\setlength{\abovecaptionskip}{-0.3cm}
\setlength{\belowcaptionskip}{-0.5cm}
\begin{center}
\includegraphics[width=1.0\linewidth]{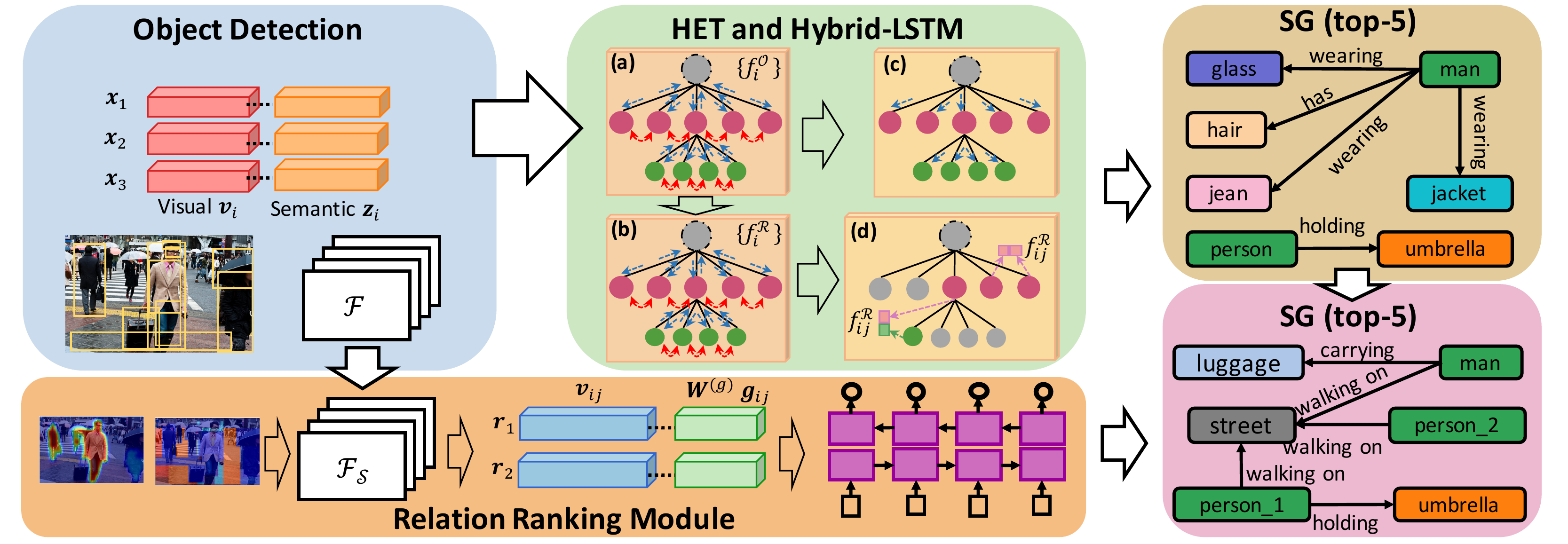}
\end{center}
   \caption{An overview of our method. An object detector is firstly applied to give support to HET construction. Then Hybrid-LSTM is leveraged to parse HET, and specifically contains 4 processes, (a) entity context encoding, (b) relation context encoding, (c) entity context decoding, and (d) relation context decoding. Finally, RRM predicts a ranking score for each triplet which further prioritizes the key relations in the scene graph.}
\label{fig:framework}
\end{figure}

\subsection{HET Construction} \label{sec:tree}

We aim to construct a hierarchical structure whose top-down levels are accord with the perceptive levels of humans' inherent scene parsing hierarchy. From a massive number of observations, it can be found that entities with larger sizes are relatively more likely to form the major events in a scene (this will be proved effective through experiments). Therefore, we arrange larger entities as close to the root of HET as possible. Each entity can be decomposed into finer entities that make up the inferior level. 

Concretely, HET is a multi-branch tree $\mathcal{T}$ with a virtual root $o_{0}$ standing for the whole image. All the entities are sorted in descending order according to their sizes and we get an orderly sequence $\{o_{i_1}, o_{i_2}, \ldots, o_{i_N}\}$. For each entity $o_{i_n}$, we consider entities with larger size, $\{o_{i_m}\}, 1 \leq m<n$, and calculate the ratio 
\begin{equation}
P_{nm}=\frac{I\left(o_{i_n}, o_{i_m}\right)}{A(o_{i_n})},
\end{equation}
where $A(\cdot)$ denotes the size of the entity and $I(\cdot,\cdot)$ is the intersection area of two entities. If $P_{nm}$ is larger than threshold $T$, $o_{i_m}$ will be a candidate parent node of $o_{i_n}$ since $o_{i_m}$ contains most part of $o_{i_n}$. If there is no candidate, the parent node of $o_{i_n}$ is set as $o_0$. If there are more than one, we further determine the parent with two alternative strategies:

\noindent{\textbf{Area-first Strategy (AFS).}} Considering that entity with a larger size has a higher probability to  contain more details or components, the candidate with the largest size is selected to be a parent node. 

\noindent{\textbf{Intersection-first Strategy (IFS).}} We compute ratio
\begin{equation}
Q_{nm}=\frac{I\left(o_{i_n}, o_{i_m}\right)}{A(o_{i_m})}.
\end{equation}
A larger $Q_{nm}$ means that $o_{i_n}$ is relatively more important to $o_{i_m}$ than to other candidates. Therefore, $o_{i_m}$ where $m=\mathop{\arg \max}_k Q_{nk}$ is chosen as parent of $o_{i_n}$. 

\subsection{Structured Context Encoding and Scene Graph Generation}\label{sec:hlstm}
The interpretability of HET implies that important relations are more likely to be seen between entities either inside a certain level or from two adjacent levels.
Therefore, both hierarchical connection  \cite{Tang_2019_CVPR} and sibling association \cite{zellers2018neural} are useful for context modeling. Our Hybrid-LSTM encoder is proposed, which consists of a bidirectional multi-branch TreeLSTM \cite{tai2015improved} (Bi-TreeLSTM) for encoding the hierarchy context, and a bidirectional chain LSTM \cite{graves2005framewise} (Bi-LSTM) for encoding the siblings context. We use two identical Hybrid-LSTM encoders to encode two types of context for each entity, one is \textbf{entity context} which helps predict the information of entity itself, and the other is \textbf{relation context} which plays a role in inferring the relation when interacting with other potential relevant entities. For brevity we only provide a detailed introduction of entity context encoding (Figure \ref{fig:framework}(a)). Specifically, the input feature $\bm{x}_i$ of each node $o_i$ is concatenation of visual feature $\bm{v}_i$ and weighted sum of semantic embedding vectors, $\bm{z}_i=\bm{W}_e^{(1)}\bm{q}_i$, where $\bm{W}_e^{(1)}$ is word embedding matrix initialized from GloVe \cite{pennington2014glove}.
For the root node $o_0$, $\bm{v}_0$ is obtained with the whole-image bounding box, while $\bm{z}_0$ is initialized randomly.

The hierarchy context (blue arrows in Figure \ref{fig:framework}(a)) is encoded as:
\begin{equation}
\bm{C} = \mathrm{BiTreeLSTM}(\{\bm{x}_i\}_{i=0}^N),
\end{equation}
where $\bm{C}=\{\bm{c}_i\}_{i=0}^N$ and each $\bm{c}_i=\left[\overrightarrow{\bm{h}_i^{\mathcal{T}}}; \overleftarrow{\bm{h}_i^\mathcal{T}}\right]$ is the concatenation of the top-down and bottom-up hidden states of Bi-TreeLSTM:
{\setlength\abovedisplayskip{1pt}
\setlength\belowdisplayskip{2pt}
\begin{subequations}
\begin{align}
\label{eq4a}
\overrightarrow{\bm{h}_i^{\mathcal{T}}} &= \mathrm{TreeLSTM}\left(\bm{x}_i, \overrightarrow{\bm{h}_p^{\mathcal{T}}}\right), \\
\overleftarrow{\bm{h}_i^{\mathcal{T}}} &= \mathrm{TreeLSTM}\left(\bm{x}_i, \left\{\overleftarrow{\bm{h}_j^{\mathcal{T}}}\Big|{j\in C(i)}\right\}\right),
\end{align}
\end{subequations}}where $C(\cdot)$ denotes the set of children nodes while subscript $p$ denotes the parent of node $i$. 

The siblings context (red arrows in Figure \ref{fig:framework}(a)) is encoded within each set of children nodes which share the same parent:
\begin{equation}
\bm{S} = \mathrm{BiLSTM}(\{\bm{x}_i\}_{i=0}^N),
\end{equation}
where $\bm{S}=\{\bm{s}_i\}_{i=0}^N$ and each $\bm{s}_i=\left[\overrightarrow{\bm{h}_i^{\mathcal{L}}}; \overleftarrow{\bm{h}_i^\mathcal{L}}\right]$ is concatenation of forward and backward hidden states of Bi-LSTM:
\begin{equation}
\overrightarrow{\bm{h}_i^{\mathcal{L}}} = \mathrm{LSTM}\left(\bm{x}_i, \overrightarrow{\bm{h}_l^{\mathcal{L}}}\right), \,\,
\overleftarrow{\bm{h}_i^{\mathcal{L}}} = \mathrm{LSTM}\left(\bm{x}_i, \overleftarrow{\bm{h}_r^{\mathcal{L}}}\right),
\end{equation}
where $l$ and $r$ stand for left and right sibling which share the same parent with $i$. We further concatenate hierarchy and siblings context to obtain the entity context, $\bm{f}^{\mathcal{O}}_i=[\bm{c}_i; \bm{s}_i]$. Missing branches or siblings are padded with zero vectors. 

The relation context is encoded (Figure \ref{fig:framework}(b)) in the same way as entity context except that the input of each node is replaced by $\{\bm{f}^{\mathcal{O}}_i\}_{i=0}^N$ . Another Hybrid-LSTM encoder is applied to get the relation context $\{\bm{f}^{\mathcal{R}}_i\}_{i=0}^N$.

To generate a scene graph, we should decode the context to obtain entity and relation information. 
In HET, a child node strongly depends on its parent, \ie, information of parent node is helpful for prediction of child node. Therefore, to predict entity information, we decode entity context in a top-down manner following Eq. \eqref{eq4a} as shown in Figure \ref{fig:framework}(c). For node $o_i$, the input $\bm{x}_i$ in Eq. \eqref{eq4a} is replaced with $[\bm{f}_i^\mathcal{O}; \bm{W}_e^{(2)}\bm{q}_p]$, where $\bm{W}_e^{(2)}$ is word embedding matrix and $\bm{q}_{p}$ is the predicted class probability vector of the parent of $o_i$. The output hidden state is fed into a softmax classifier and bounding box regressor to predict entity information of $o_i$. To predict the predicate $p_{ij}$ between $o_i$ and $o_j$, we feed $\bm{f}_{ij}^\mathcal{R}=[\bm{f}_i^\mathcal{R}; \bm{f}_j^\mathcal{R}]$ to an MLP classifier (Figure \ref{fig:framework}(d)). As a result, a scene graph is generated, and for each triplet containing subject $o_i$, object $o_j$ and predicate $p_{ij}$, we obtain their scalar scores $s_i$, $s_j$, and $s_{ij}$. 

\subsection{Relation Ranking Module}\label{sec:RRM}
So far, we obtain the hierarchical scene graph based on HET. As we collect the key relation annotations (Section \ref{subsec:data}), we intend to further maximize the performance on mining key relations with supervised information, and explore the advantages brought by HET. 
Consequently, we design a Relation Ranking Module (RRM) to prioritize key relations. As analyzed in \textbf{Related Works}, regions of humans' interest can be tracked under the guidance of \textit{visual saliency} although they do not always form the major events that humans want to convey. Besides, the \textit{size}, which guides HET construction, not only is an important reference for estimating the perceptive level of entities, but also is found helpful to rectify some misleadings in humans' subjective assessment on the importance of relations (see the Appendix). Therefore, we propose to learn to capture humans' subjective assessment on the importance of relations under the guidance of visual saliency and entity size information. 

We firstly employ DSS \cite{HouCvpr2017Dss} to predict the pixel-wise saliency map (\textbf{SM}) $\mathcal{S}$ for each image. To effectively collect entity size information, we propose a pixel-wise area map (\textbf{AM}) $\mathcal{A}$. Given the image $\mathcal{I}$ and its detected $N$ entities $\{o_i\}_{i=1}^N$ with bounding boxes $\{\bm{b}_i\}_{i=1}^N$ (specially $o_0$ and $\bm{b}_0$ for the whole image), the value $a_{xy}$ of each position $(x, y)$ on $\mathcal{A}$ is defined as the minimum normalized size of entities which cover $(x, y)$:
{\setlength\abovedisplayskip{1pt}
\setlength\belowdisplayskip{2pt}
\begin{equation}
a_{xy} = \left\{
\begin{aligned}
& \min\left\{ \frac{A(o_i)}{A(o_0)} \Bigg| i\in\mathcal{X}\right\}, \mathrm{if}\,\, \mathcal{X}\neq \emptyset \\
& 0, \mathrm{otherwise},
\end{aligned}\right.
\end{equation}}where $\mathcal{X}=\{i|(x,y)\in \bm{b}_i, 0<i\leq N\}$. The sizes of both $\mathcal{S}$ and $\mathcal{A}$ are the same as that of input image $\mathcal{I}$. We apply adaptive average pooling ($\mathrm{AAP}(\cdot)$) to smooth and down-sample these two maps to align with the shape of conv5 feature map $\mathcal{F}$ from Faster-RCNN, and obtain the attention embedded feature map $\mathcal{F}_S$:
{\setlength\abovedisplayskip{1pt}
\setlength\belowdisplayskip{2pt}
\begin{equation}\label{eq:saliency}
\mathcal{F}_S = \mathcal{F} \odot (\mathrm{AAP}(\mathcal{S}) + \mathrm{AAP}(\mathcal{A})),
\end{equation}}where $\odot$ is the Hadamard product.

We predict a score for each triplet to adjust their rankings. The input contains visual representation for a triplet, $\bm{v}_{ij} \in \mathbb{R}^{4096}$, which is obtained by RoI Pooling on $\mathcal{F}_S$. Besides, the geometric information is also an auxiliary cue for estimating the importance. For a triplet containing subject box $\bm{b}_i$ and object box $\bm{b}_j$, the geometric feature $\bm{g}_{ij}$ is defined as a 6-dimensional vector following \cite{Kim_2019_CVPR}:
{\setlength\abovedisplayskip{1pt}
\setlength\belowdisplayskip{2pt}
\begin{equation}
\bm{g}_{ij} =\!\!\left[ \frac{x_j - x_i}{\sqrt{w_i h_i}},\! \frac{y_j-y_i}{\sqrt{w_i h_i}}, \!\sqrt{\frac{w_j h_j}{w_i h_i}},\! \frac{w_i}{h_i},\! \frac{w_j}{h_j}, \!\frac{\bm{b}_i\cap \bm{b}_j}{\bm{b}_i \cup \bm{b}_j}\right]\!\!,
\end{equation}}which is projected to a 256-dimensional vector and concatenated with $\bm{v}_{ij}$, resulting in the final representation for a relation $\bm{r}_{ij}=[\bm{v}_{ij}; \bm{W}^{(g)}\bm{g}_{ij}]$ where $\bm{W}^{(g)} \in \mathbb{R}^{256\times 6}$ is projection matrix. 
Then we use a bi-directional LSTM to encode global context among all the triplets so that ranking score of each triplet can be reasonably adjusted considering scores of other triplets.  Concretely, the ranking score $t_{ij}$ for a pair $(o_i$, $o_j)$ is achieved as:
\begin{align}
\label{eq:rrm1} \{\bm{h}^{\mathcal{R}}_{ij}\} &= \mathrm{BiLSTM}\left(\{\bm{r}_{ij}\}\right), \\
\label{eq:rrm2} t_{ij} &= \bm{W}^{(r)}_2\mathrm{ReLU}(\bm{W}^{(r)}_1\bm{h}^{\mathcal{R}}_{ij}).
\end{align}
$\bm{W}^{(r)}_1$ and $\bm{W}^{(r)}_2$ are weights of two fully connected layers. The ranking score is fused with classification scores so that both the confidences of three components of a triplet and ranking priority are considered, resulting in the final ranking confidence $\phi_{ij} = s_i\cdot s_j\cdot s_{ij}\cdot t_{ij}$, which is used for re-ranking the relations. 

\subsection{Loss Function}
We adopt the cross-entropy loss for optimizing Hybrid-LSTM networks. Let $e'$ and $l'$ denote the predicted label of entity and predicate respectively, $e$ and $l$ denote the ground truth labels. The loss is defined as:
{\setlength\abovedisplayskip{1pt}
\setlength\belowdisplayskip{1pt}
\begin{equation}
\begin{split}
\mathcal{L}_{CE} = \mathcal{L}_{entity} +  \mathcal{L}_{relation} 
= -\frac{1}{Z_1} \sum_{i} e'_i \log (e_i) -\frac{1}{Z_2} \sum_{i}\sum_{j\neq i} l'_{ij} \log (l_{ij}).
\end{split}
\end{equation}}

When the RRM is applied, the final loss function is the sum of $\mathcal{L}_{CE}$ and ranking loss $\mathcal{L}(\mathcal{K}, \mathcal{N})$, which is used to maximize the margin between the ranking confidences of key relations and those of secondary ones:
{\setlength\abovedisplayskip{1pt}
\setlength\belowdisplayskip{2pt}
\begin{equation}
\begin{split}
\mathcal{L}_{Final} = \mathcal{L}_{CE} + \mathcal{L}(\mathcal{K}, \mathcal{N})
=\mathcal{L}_{CE} +\frac{1}{Z_3} \sum_{r\in \mathcal{K}, r'\in \mathcal{N}}\, \max (0, \gamma - \phi_r + \phi_{r'}),
\end{split}
\end{equation}}where $\gamma$ denotes margin parameter, $\mathcal{K}$ and $\mathcal{N}$ stand for the set of key and secondary relations, $r$ and $r'$ are relations sampled from $\mathcal{K}$ and $\mathcal{N}$ with ranking confidences $\phi_{r}$ and $\phi_{r'}$. $Z_1$, $Z_2$, and $Z_3$ are  normalization factors. 

\section{Experimental Evaluation}
\subsection{Dataset, Evaluation and Settings}\label{subsec:data}

\noindent{\textbf{VRD}} \cite{lu2016visual}, is the benchmarking dataset for visual relationship detection task, which contains 4,000/1,000 training/test images and covers 100 object categories and 70 predicate categories.

\noindent{\textbf{Visual Genome (VG),}} is a large-scale dataset with rich annotations of objects, attributes, dense captions and  pairwise relationships, containing 75,651/32,422 training/test images. We adopt the most widely used version of VG, namely VG150 \cite{xu2017scene}, which covers 150 object categories and 50 predicate categories.

\begin{figure}[t]
\setlength{\abovecaptionskip}{-0.2cm}
\setlength{\belowcaptionskip}{-0.5cm}
\begin{center}
\includegraphics[width=1.0\linewidth]{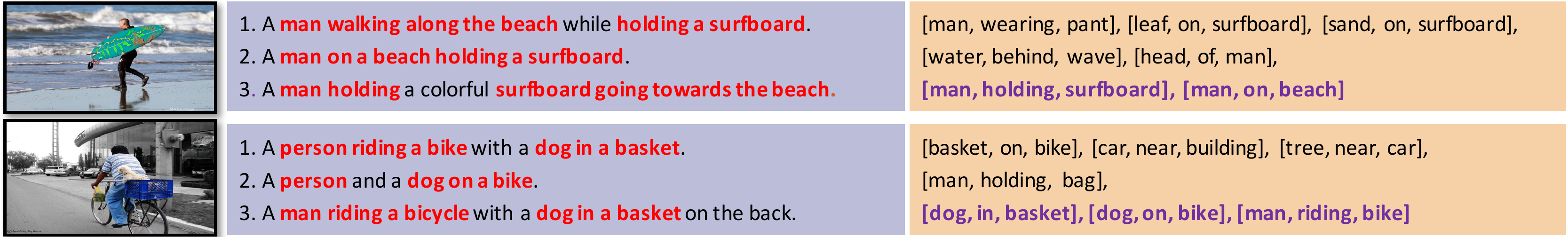}
\end{center}
   \caption{Examples in VG-KR dataset. Each image is shown with 3 captions and ground truth relations. Purple triplets are key ones while others are secondary.}
\label{fig:vg_kr}
\end{figure}

\noindent{\textbf{VG200 and VG-KR.}} We intend to collect the indicative annotations of key relations based on VG. Inspired by the finding illustrated in \textbf{Related Works}, we associate the relation triplets referred in caption annotations in MSCOCO \cite{lin2014microsoft} with those from VG. We give several examples in Figure \ref{fig:vg_kr}. The details of our processing and more statistics are provided in the Appendix.


\noindent{\textbf{Evaluation, Settings, and Implementation Details.}} For conventional SGG following triplet-match rule (only if three components of a triplet match the ground truth will it be a correct one), we adopt three universal protocols \cite{xu2017scene}: PREDCLS, SGCLS, and SGGEN. All protocols use Recall@K (R@K=20, 50, 100) as a metric. When evaluating key relation prediction, there are some variations. First, we only evaluate with PREDCLS and SGCLS protocols to eliminate the interference of errors from object detector, and add a tuple-match rule (only the subject and object are required to match the ground truth) to investigate the ability to find proper pairs. Second, we introduce a new metric, \textbf{Key Relation Recall (kR@K)}, which computes recall rate on key relations. As the number of key relations is usually less than 5 (see the Appendix), the K in kR@K is set to 1 and 5. When evaluating on VRD, we use RELDET and PHRDET \cite{yu2017visual}, and report R@50 and R@100 at 1, 10, and 70 predicates per related pair. The details about the hyperparameters settings and implementation are provided in the Appendix. 

%


\setlength{\tabcolsep}{2pt}
\begin{table}
\setlength{\belowcaptionskip}{-0.3cm}
\renewcommand\arraystretch{0.7}
\begin{center}
\caption{Results table (\%) on VG150 and VG200. The results of the full version of our method are highlighted.}
\label{tab:vg150-sg}
\resizebox{\textwidth}{!}{
\begin{tabular}{@{}llccccccccc@{}}\toprule
& & \multicolumn{3}{c}{SGGEN} & \multicolumn{3}{c}{SGCLS} & \multicolumn{3}{c}{PREDCLS} \\\cmidrule(lr){3-5}\cmidrule(lr){6-8}\cmidrule(l){9-11}
 & R@ & 20 & 50 & 100 & 20 & 50 & 100 & 20 & 50 & 100 \\\hline
\multirow{12}*{\rotatebox{90}{VG150}}& VRD \cite{lu2016visual} & - & 0.3 & 0.5 & - & 11.8 & 14.1 & - & 27.9 & 35.0 \\
& IMP \cite{xu2017scene} & - & 3.4 & 4.2 & - & 21.7 & 24.4 & - & 44.8 & 53.0 \\
& IMP$\dag$ \cite{xu2017scene,zellers2018neural} & 14.6 & 20.7 & 24.5 & 31.7 & 34.6 & 35.4 & 52.7 & 59.3 & 61.3 \\

& Graph-RCNN \cite{yang2018graph} & - & 11.4 & 13.7 & - & 29.6 & 31.6 & - & 54.2 & 59.1\\

& MemNet \cite{Wang_2019_CVPR} & 7.7 & 11.4 & 13.9 & 23.3 & 27.8 & 29.5 & 42.1 & 53.2 & 57.9 \\
& MOTIFS \cite{zellers2018neural} & 21.4 & 27.2 & 30.3 & 32.9 & 35.8 & 36.5 & 58.5 & 65.2 & 67.1  \\
& KERN \cite{Chen_2019_CVPR} & - & 27.1 & 29.8 & - & 36.7 & 37.4 & - & 65.8 & 67.6 \\
& VCTree-SL \cite{Tang_2019_CVPR} &21.7 & 27.7 & 31.1 & 35.0 & 37.9 & 38.6 & 59.8 & 66.2 & 67.9 \\
\cmidrule(){2-11}
& HetH-AFS & 21.2 & 27.1 & 30.5 & 33.7 & 36.6 & 37.3 & 58.1 & 64.7 & 66.6\\
& HetH w/o chain & 21.5 & 27.4 & 30.7 & 32.9 & 35.9 & 36.7 & 57.5 & 64.5 & 66.5\\
& HetH & \textbf{21.6} & \textbf{27.5} & \textbf{30.9} & \textbf{33.8} & \textbf{36.6} & \textbf{37.3}  & \textbf{59.8} & \textbf{66.3} & \textbf{68.1}\\
\cmidrule(){1-11}
\multirow{3}*{\rotatebox{90}{\small{VG200}}}
& MOTIFS \cite{zellers2018neural} & 15.2 & 19.9 & 22.8 & 24.5 & 26.7 & 27.4 & 52.5 & 59.0 & 61.0  \\
& VCTree-SL \cite{Tang_2019_CVPR} & 14.7 & 19.5 & 22.5 & 24.2 & 26.5 & 27.1 & 51.9 & 58.4 & 60.3 \\
& HetH & \textbf{15.7} & \textbf{20.4} & \textbf{23.4} & \textbf{25.0} & \textbf{27.2} & \textbf{27.8}  & \textbf{53.6} & \textbf{60.1} & \textbf{61.8} \\
\bottomrule
\end{tabular}}
\end{center}
\end{table}
\setlength{\tabcolsep}{2.0pt}

\setlength{\tabcolsep}{8.0pt}
\begin{table}
\setlength{\belowcaptionskip}{-1.5cm}
\renewcommand\arraystretch{0.87}
\begin{center}
\caption{Results table (\%) of key relation prediction on VG-KR. }
\label{tab:vg-kr}
\resizebox{\textwidth}{!}{
\begin{tabular}{@{}lcccccccc@{}}\toprule
 & \multicolumn{4}{c}{Triplet Match} & \multicolumn{4}{c}{Tuple Match} \\\cmidrule(lr){2-5}\cmidrule(lr){6-9}
 & \multicolumn{2}{c}{SGCLS} & \multicolumn{2}{c}{PREDCLS} & \multicolumn{2}{c}{SGCLS} & \multicolumn{2}{c}{PREDCLS} \\
\cmidrule(lr){2-3}\cmidrule(lr){4-5}
\cmidrule(lr){6-7}\cmidrule(lr){8-9}
  kR@ & 1 & 5 & 1 & 5 &  1 & 5 & 1 & 5\\\hline
  VCTree-SL & 5.7 & 14.2 & 11.4 & 30.2 & 8.4 & 22.2 & 16.1 & 46.4 \\ 
  MOTIFS & 5.9 & 14.5 & 11.3 & 30.0 & 8.5 & 21.8 & 16.0 & 46.2 \\  
  HetH & \textbf{6.1} & \textbf{15.1} & \textbf{11.6} & \textbf{30.4} & \textbf{8.6} & \textbf{22.7} & \textbf{16.4} & \textbf{47.1} \\\hline
  MOTIFS-RRM & 8.6 & 16.4 & 16.7 & 33.8 & 13.8 & 26.3 & 27.9 & 57.1 \\
  HetH-RRM & \textbf{9.2} & \textbf{17.1} & \textbf{17.5} & \textbf{35.0} & \textbf{14.6} & \textbf{27.3} & \textbf{28.9} & \textbf{59.1} \\\hline
  RRM-Base & 8.4 & 16.8 & 16.2 & 33.7 & 13.4 & 26.8 & 26.6 & 57.2 \\
  RRM-SM & 9.0 & 16.9 & 17.2 & 34.5 & 14.3 & 27.1 & 28.6 & 58.7 \\
  RRM-AM & 8.9 & 16.9 & 16.9 & 34.4 & 14.1 & 27.0 & 28.1 & 58.2 \\ 
\bottomrule
\end{tabular}}
\end{center}
\end{table}

\setlength{\tabcolsep}{8.0pt}
\begin{table}
\renewcommand\arraystretch{0.9}
\begin{center}
\caption{Results table (\%) on VRD. }
\label{tab:vrd}
\resizebox{\textwidth}{!}{
\begin{tabular}{@{}lcccccccccccc@{}}\toprule
 & \multicolumn{6}{c}{RELDET} & \multicolumn{6}{c}{PHRDET} \\\cmidrule(lr){2-7}\cmidrule(lr){8-13}
 & \multicolumn{2}{c}{k=1} & \multicolumn{2}{c}{k=10} & \multicolumn{2}{c}{k=70} & \multicolumn{2}{c}{k=1} & \multicolumn{2}{c}{k=10} & \multicolumn{2}{c}{k=70}\\
\cmidrule(lr){2-3}\cmidrule(lr){4-5}
\cmidrule(lr){6-7}\cmidrule(lr){8-9}
\cmidrule(lr){10-11}\cmidrule(lr){12-13}
  R@ & 50 & 100 & 50 & 100 & 50 & 100 & 50 & 100 & 50 & 100 & 50 & 100 \\\hline
  ViP \cite{li2017vip} & 17.32 & 20.01 & - & - & - & - & 22.78 & 27.91 & - & - & - & - \\
  VRL \cite{liang2017deep} & 18.19 & 20.79 & - & - & - & - & 21.37 & 22.60 & - & - & - & - \\
  KL-Dist \cite{yu2017visual} & 19.17 & 21.34 & 22.56 & 29.89 & 22.68 & 31.89 & 23.14 & 24.03 & 26.47 & 29.76 & 26.32 & 29.43 \\
  Zoom-Net \cite{yin2018zoom} & 18.92 & 21.41 & - & - & 21.37 & 27.30 & 24.82 & 28.09 & - & - & 29.05 & 37.34 \\
  RelDN-$L_0$ \cite{Zhang2019graphical} & 24.30 &27.91 & 26.67 & 32.55 & 26.67 & 32.55 & 31.09 &36.42 & 33.29 & 41.25 & 33.29 & 41.25 \\
  RelDN \cite{Zhang2019graphical} & 25.29 & 28.62 & 28.15 & 33.91 & 28.15 & 33.91 & 31.34 & 36.42 &34.45 & 42.12 & 34.45 & 42.12\\\hline

  HetH & 22.42 & 24.88 & 26.88 & 31.69 & 26.88 & 31.81 & 30.69 & 35.59 & \textbf{35.47} & \textbf{42.94} & \textbf{35.47} & \textbf{43.05}
   \\ 
\bottomrule
\end{tabular}}
\end{center}
\end{table}

\subsection{Ablation Studies}
Ablation studies are separated into two sections. The first part is to explore some variants of HET construction. We conduct these experiments on VG150. The complete version of our model is \textbf{HetH}, which is configured with IFS and Hybrid-LSTM. 
The second part is an investigation into the usage of SM and AM in RRM. Experiments are carried out on VG-KR. The complete version is \textbf{HetH-RRM}, whose implementation follows Eq. \eqref{eq:saliency}. 

\textbf{Ablation study on HET construction. }
We firstly compare \textbf{AFS} and \textbf{IFS} for determining the parent node. Then we investigate the effectiveness of the chain LSTM encoder in Hybrid-LSTM. The ablative models mentioned above are shown in Table \ref{tab:vg150-sg} as \textbf{HetH-AFS} (\ie replace IFS by AFS), and \textbf{HetH w/o chain}. We observe that using IFS together with Hybrid-LSTM encoder has the best performances, which indicates that HET would be more reasonable using IFS. It's noteworthy that if the Bi-TreeLSTM encoder is abandoned, the Hybrid-LSTM encoder would almost degenerate to MOTIFS. Therefore, through comparisons between HetH and MOTIFS, HetH and HetH w/o chain, it implies that both hierarchy and siblings context should be encoded in HET. 

\textbf{Ablation study on RRM.} In order to explore the effectiveness of saliency and size, we ablate HetH-RRM with the following baselines: (1) \textbf{RRM-Base:} $\bm{v}_{ij}$ is extracted from $\mathcal{F}$ rather than $\mathcal{F}_S$, (2) \textbf{RRM-SM:} only $\mathcal{S}$ is used, and (3) \textbf{RRM-AM:} only $\mathcal{A}$ is used. Results in Table \ref{tab:vg-kr} suggest that both saliency and size information indeed contributes to discovering key relations, and the effect of saliency is slightly better than that of the size. The hybrid version achieves the highest performances. From the following qualitative analysis, we can see that with the guidance of saliency and rectification effect of size, RRM further shifts the model's attention to key relations significantly.

\subsection{Comparisons with State-of-the-Arts}
For scene graph generation, we compare our \textbf{HetH} with the following state-of-the-art methods: \textbf{VRD} \cite{lu2016visual} and \textbf{KERN} \cite{Chen_2019_CVPR} use knowledge from language or statistical correlations. \textbf{IMP} \cite{xu2017scene}, \textbf{Graph-RCNN} \cite{yang2018graph}, \textbf{MemNet} \cite{Wang_2019_CVPR}, \textbf{MOTIFS} \cite{zellers2018neural} and \textbf{VCTree-SL} \cite{Tang_2019_CVPR} mainly devise various message passing methods for improving graph representations. 
For key relation prediction, we mainly evaluate two latest works, MOTIFS and VCTree-SL on VG-KR. Besides, we further incorporate RRM to MOTIFS, namely \textbf{MOTIFS-RRM}, to explore the transferability of RRM. Results are shown in Table \ref{tab:vg150-sg} and \ref{tab:vg-kr}. We give statistical significance of the results in the Appendix.  

\begin{figure}[t]
\setlength{\abovecaptionskip}{-0.2cm}
\setlength{\belowcaptionskip}{-0.3cm}
\begin{center}
\includegraphics[width=1.0\linewidth]{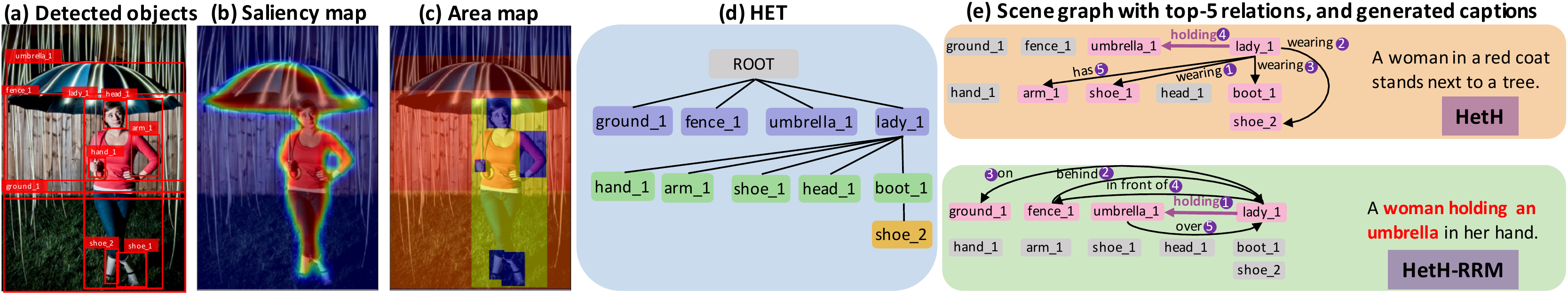}
\end{center}
   \caption{Qualitative Results of HetH and HetH-RRM. In (e), the pink entities are involved in top-5 relations, and the purple arrows are key relations matched with ground truth. The purple numeric tags next to the relations are the rankings, and \lq\lq 1\rq\rq\,\, means that the relation gets the highest score.}
\label{fig:results}
\end{figure}

\textbf{Quantitative Analysis.} 
In Table \ref{tab:vg150-sg}, when evaluated on \textbf{VG150}, HetH dominantly surpasses most methods. Compared to MOTIFS and VCTree-SL, HetH using multi-branch tree structure outperforms MOTIFS and yields comparable recall rate with VCTree-SL which uses a binary tree structure. It indicates that hierarchical structure is superior to plain one in terms of modeling context. We observe that HetH achieves better performances compared to VCTree-SL under PREDCLS protocol, while there exists a slight gap under SGCLS and SGGEN protocols. This is mainly because our tree structure is generated with artificial rules and some incorrect subtrees inevitably emerge due to occlusion in 2D images, while VCTree-SL dynamically adjusts its structure in pursuit of higher performances. Under SGCLS and SGGEN protocols in which object information is fragmentary, it is difficult for HetH to rectify the context encoded from wrong structures.  However, we argue that our interpretable and natural multi-branch tree structure is also adaptive to the situation when there is an increment of object and relation categories but fewer data. It can be seen from evaluation results on \textbf{VG200} that the HetH outperforms MOTIFS by 0.67 mean points and VCTree-SL by 1.1 mean points. On the contrary, in this case, the data are insufficient for dynamic structure optimization. 

As SGG task is highly related to VRD task, we apply HetH on \textbf{VRD} and the comparison results are shown in Table \ref{tab:vrd}. Both the  HetH and RelDN \cite{Zhang2019graphical} use pre-trained weights on MSCOCO, while only \cite{yin2018zoom} states that they use ImageNet pre-trained weights and others remain unknown. It's shown that our method yields competitive results and even surpasses state-of-the-arts under some metrics. 

When it comes to key relation prediction, we directly evaluate HetH, MOTIFS, and VCTree-SL on \textbf{VG-KR}. As shown in Table \ref{tab:vg-kr}, HetH substantially performs better than other two competitors, suggesting that the structure of HET provides hints for judging the importance of relations, and parsing the structured information in HET indeed capture humans' perceptive habits. 

In pursuit of ultimate performances on mining key relations, we jointly optimize the HetH with RRM under the supervision of key relation annotations in VG-KR. From Table \ref{tab:vg-kr}, both HetH-RRM and MOTIFS-RRM achieve significant gains, and HetH-RRM is better than MOTIFS-RRM, which proves the superiority of HetH again, and shows excellent transferability of RRM.

\textbf{Qualitative Analysis.} We visualize intermediate results in Figure \ref{fig:results}(a-d). HET is well constructed and close to human's analyzing process. In the area map, regions of \textit{arm}, \textit{hand}, and \textit{foot} get small weights because of their small sizes. Actually, relations like $\langle$\textit{lady}, \textit{has}, \textit{arm}$\rangle$ are indeed trivial. As a result, RRM suppresses these relations. More cases are provided in the Appendix.

\begin{figure}[htb] 
\setlength{\belowcaptionskip}{-1.0cm}
  \begin{minipage}[b]{0.58\textwidth} 
    \centering 
    \subfloat[]{\includegraphics[width=0.5\textwidth]{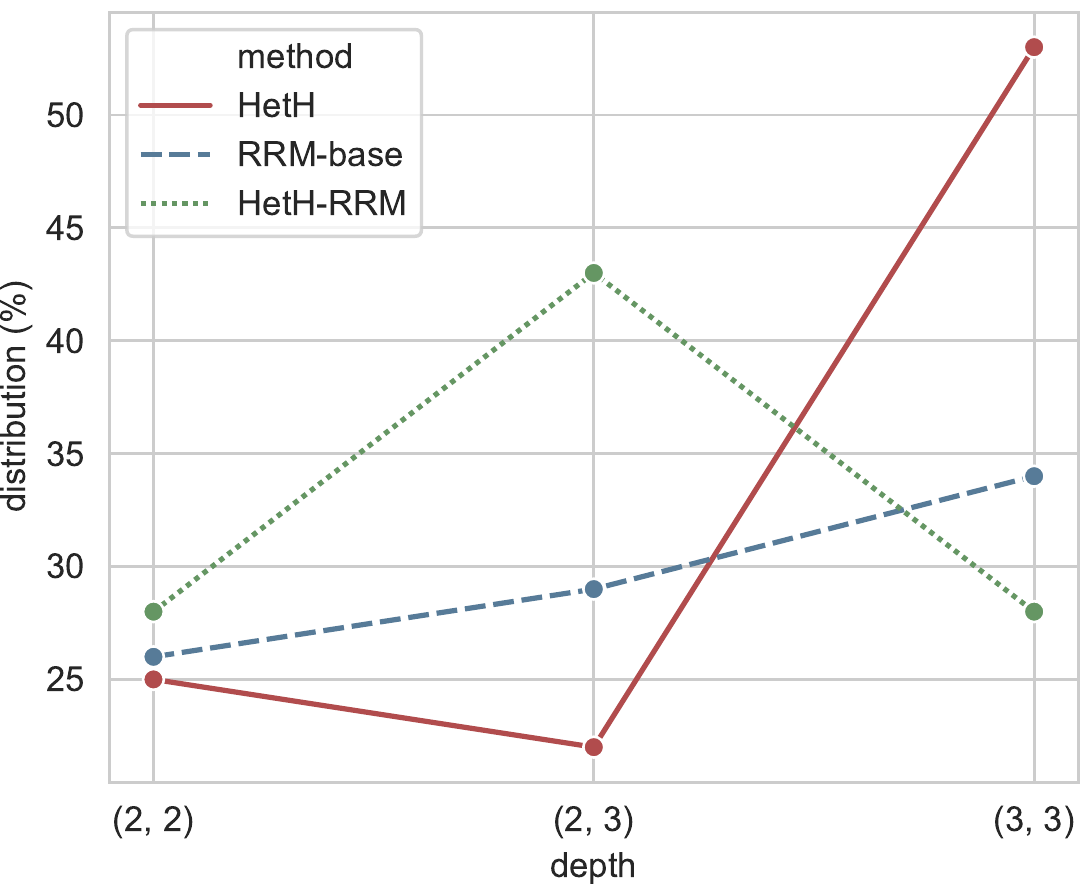}} 
    \subfloat[]{
    \includegraphics[width=0.5\textwidth]{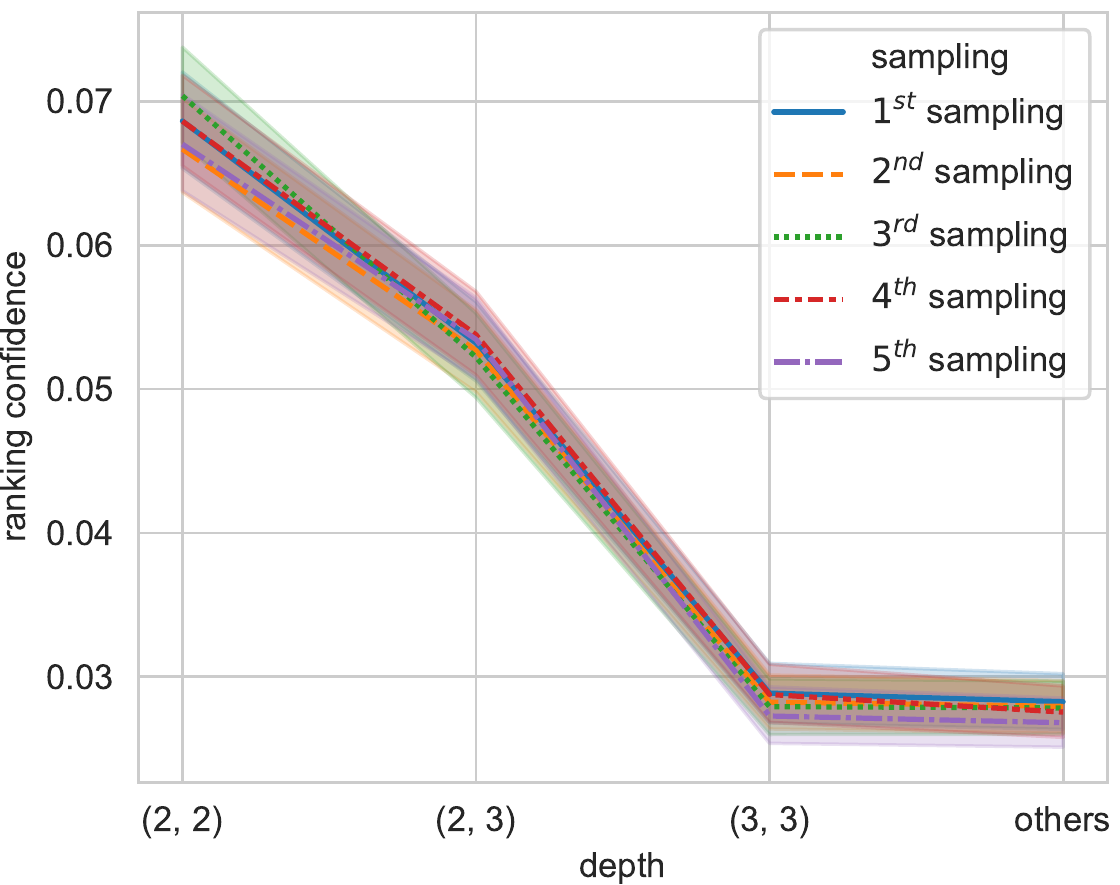}}
    \caption{(a) Depth distribution of top-5 predicted relations. (b) The ranking confidence of relations from different depths obtained from RRM-base. Sampling is repeated five times. } 
    \label{fig:alsdepth} 
  \end{minipage}\,\,\,
   \begin{minipage}[b]{0.4\textwidth} 
    \centering
    \resizebox{\textwidth}{!}{
    \begin{tabular}{|c|c|c|} \hline 
      Strtg. & Metric & HetH-RRM \\ \hline\hline 
      \multirow{3}*{EP} & kR@1 & \textbf{17.5} \\ 
      & kR@5 & \textbf{35.0} \\ 
      & speed & 0.22  \\\hline
      \multirow{3}*{SP} & kR@1 & 15.8 \\
      & kR@5 & 31.2 \\
      & speed & \textbf{0.18}\\\hline 
    \end{tabular}} 
    \caption{Comparison between EP and SP. The inference speed (seconds/image) is evaluated with a single TITAN Xp GPU).} 
    \label{tab:alstree} 
  \end{minipage}
\end{figure}

\begin{figure}[t]
\setlength{\abovecaptionskip}{-0.2cm}
\setlength{\belowcaptionskip}{-0.3cm}
\begin{center}
\includegraphics[width=1.0\linewidth]{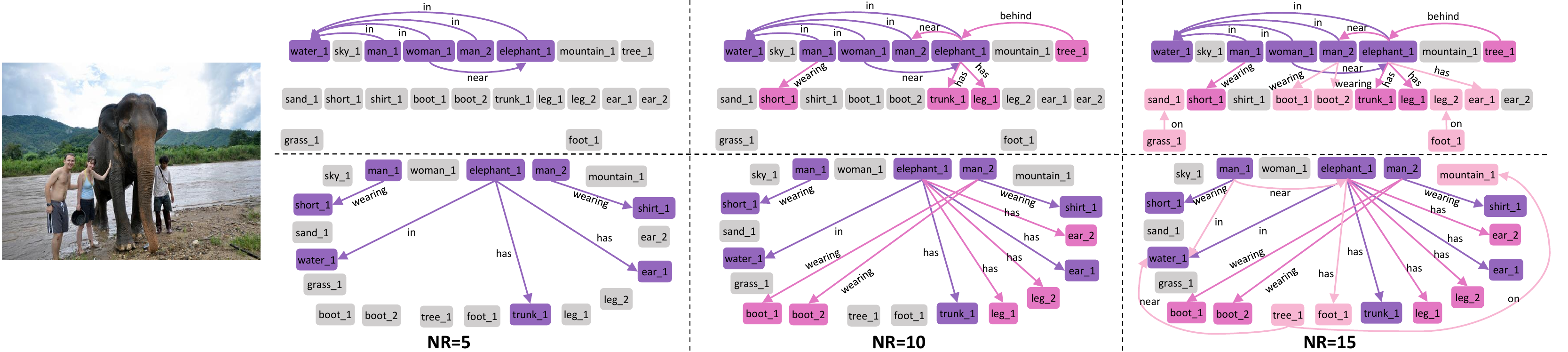}
\end{center}
   \caption{As the quota of top relations (NR) increases, scene graphs dynamically enlarge. The newly involved entities and relations are shown in a new color. Results in first row and  second row are from HetH-RRM and MOTIFS respectively.}
\label{fig:varsg}
\end{figure}

\subsection{Analyses about HET}

We conduct additional experiments to validate whether HET has a potential to reveal humans' perceptive habits. As shown in Figure \ref{fig:alsdepth}(a), we compare the \textbf{depth distribution} of top-5 predicted relations (represented by tuple $(d_{o_i}, d_{o_j})$ consisting of the depths of two entities, and the depth of root is defined as 1.) of HetH, RRM-base and HetH-RRM. After applying RRM, there is a significant increment on the ratio of depth tuples (2, 2) and (2, 3), and a drop on (3, 3). This phenomenon is also observed in Figure \ref{fig:results}(e). Previous experiments have proved that RRM obviously upgrades the rankings of key relations. In other words, relations which are closer to the root of HET are regarded as key ones by RRM. We also analyze the ranking confidence ($\phi$) of relations from different depths with the RRM-base model (to eliminate the confounding effect caused by AAP information). We sample 10,000 predicted relation triplets from each depth five times. In Figure \ref{fig:alsdepth}(b), the ranking confidence decreases as the depth increases. Therefore, different levels of HET indeed indicate different perceptive importance of relations. This characteristic makes it possible to reasonably adjust the scale of a scene graph. As shown in the first row in Figure \ref{fig:varsg}, hierarchical scene graph from our HetH-RRM enlarges in a top-down manner as the quota of top relations increases, while the ordinary scene graph in the second row enlarges itself aimlessly. If we want to limit the scale of a scene graph but keep its ability to sketch image gist as far as possible, it is feasible for our hierarchical scene graph since we just need to cut off some secondary branches of HET, but is difficult to realize in an ordinary scene graph.

Besides, different from traditional \textbf{Exhausted Prediction} (\textbf{EP}, predict relation for every pair of entities) during inference stage, we adopt a novel \textbf{Structured Prediction} (\textbf{SP}) strategy, in which we only predict relations between parent and children nodes, and any two sibling nodes that share the same parent. In Figure \ref{tab:alstree}, we compare the performances and inference speed between EP and SP for HetH-RRM. Despite the slight gap in terms of performances, the interpretability of connections in HET makes SP feasible to take a further step towards efficient inference, getting rid of the $O(N^2)$ complexity \cite{li2018factorizable} of EP. Further researches need to be conducted to balance performance and efficiency.

\setlength{\tabcolsep}{2pt}
\begin{table}
\renewcommand\arraystretch{0.9}
\begin{center}
\caption{Results of image captioning on VG-KR.}
\label{tab:captions}
\resizebox{\textwidth}{!}{
\begin{tabular}{@{}clcccccccc@{}}\toprule
  Num. & Model & B@1 & B@2 & B@3 & B@4 &  ROUGE-L & CIDEr & SPICE & Avg. Growth\\\hline
  all & GCN-LSTM & 72.0 & 54.7 & 40.5 & 30.0 & 52.9 & 91.1 & 18.1 \\\hline
  \multirow{3}*{20} 
  & HetH-Freq & 73.1 & 55.7 & 41.0 & 30.1 & 53.5 & 94.0 & 18.8\\
  & HetH & 74.9 & \textbf{58.4} & \textbf{43.9} & \textbf{32.8} & 54.9 & 101.7 & 19.8 & \multirow{2}*{0.06}\\
  & HetH-RRM & \textbf{75.0} & 58.2 & 43.7 & 32.7 & \textbf{55.1} & \textbf{102.2} & \textbf{19.9} \\\hline
  \multirow{3}*{5} 
  & HetH-Freq & 70.7 & 53.2 & 38.6 & 28.0 & 51.7 & 84.4 & 17.2\\
  & HetH & 72.5 & 55.4 & 41.2 & 30.5 & 53.1 & 92.6 & 18.5 & \multirow{2}*{1.57}\\
  & HetH-RRM & \textbf{73.7} & \textbf{56.7} & \textbf{42.3} & \textbf{31.5} & \textbf{54.0} & \textbf{97.5} & \textbf{19.1} \\\hline
  \multirow{3}*{2} 
  & HetH-Freq & 68.1 & 50.8 & 36.8 & 26.5 & 50.2 & 76.5 & 15.5\\
  & HetH & 70.8 & 53.4 & 39.2 & 28.7 & 51.8 & 86.4 & 17.6 & \multirow{2}*{2.10}\\
  & HetH-RRM & \textbf{72.3} & \textbf{55.2} & \textbf{41.0} & \textbf{30.4} & \textbf{53.1} & \textbf{92.2} & \textbf{18.4} \\
\bottomrule
\end{tabular}}
\end{center}
\end{table}

\section{Experiments on Image Captioning}
Do key relations really make sense? We conduct experiments on one of the downstream tasks of SGG, \ie, image captioning, to verify it. \footnote[2]{We briefly introduce here and details are provided in Appendix.}  

Experiments are conducted on VG-KR since it has caption annotations from MSCOCO. To generate captions, we select different numbers of predicted top relations and feed them into the LSTM backend following \cite{yao2018exploring}. We reimplement the complete \textbf{GCN-LSTM} \cite{yao2018exploring} model and evaluate it on VG-KR since it's one of the state-of-the-art methods and is most related to us. As shown in Table \ref{tab:captions}, our simple frequency baseline, \textbf{HetH-Freq} (the rankings of relations are accord with their frequency in training data), with 20 top relations input, outperforms GCN-LSTM because GCN-LSTM conducts graph convolution using relations as edges, which is not as effective as our method in terms of making full use of relation information.  
After applying RRM, there is consistent performance improvement on overall metrics. This improvement is more and more significant as the number of input top relations reduces. It's reasonable since the impact of RRM centers at top relations. It suggests that our model provides more essential content with as few relations as possible, which contributes to efficiency improvement. The captions presented in Figure \ref{fig:results}(e) shows that key relations are more helpful for generating a description that highly fits the major events in an image.

\section{Conclusion}
We propose a new scene graph modeling formulation and make an attempt to push the study of SGG towards the target of practicability and rationalization. 
We generate a human-mimetic hierarchical scene graph inspired by humans' scene parsing procedure, and further prioritize the key relations as far as possible. Based on HET, a hierarchcal scene graph is generated with the assistance of our Hybrid-LSTM. Moreover, RRM is devised to recall more key relations. Experiments show outstanding performances of our method on traditional scene graph generation and key relation prediction tasks. Besides, experiments on image captioning prove that key relations are not just for appreciating, but indeed play a crucial role in higher-level downstream tasks. 

\clearpage

\appendix
\renewcommand{\appendixname}{Appendix~\Alph{section}}

\section{Detailed Explanation about Motivation}\label{sec:mot}

As illustrated in the main paper, it's notable that the visually salient objects are related but not completely equal to objects involved in image gist. According to findings in \cite{He_2019_ICCV}, objects referred in a description (\ie, objects that humans think important and should form the major events/image gist) are almost visually salient and  reveal where humans gaze, but what humans fixate (\ie, visually salient objects) are not always what they want to convey at first. In Figure \ref{fig:mot}, we provide some examples to show that this is a common phenomenon. \Eg, the \textit{red clothes}, the \textit{Spring Festival couplets}, and the \textit{black doors of the washing machines} (mentioned from left to right), are visually salient due to their high contrast to the context or center position. However, some of them do not form the major events. For example, in the 2\textsuperscript{nd} image, the first glance description would be \lq\lq There stands a house on the side of the road\rq\rq. Then humans may be interested in the eyecatching \textit{Spring Festival couplets}.

Besides, we are inspired by these observations. There naturally exists a hierarchical structure about humans' perception preference. Objects with relatively large size which fulfill the scene generally form the major events. It supports us to construct HET with the method introduced in the main paper. We aim at constructing HET whose levels reflect the perception priority level rather than the object saliency. The experiments show that our method for constructing HET has achieved this goal.

\begin{figure}[htb]
\begin{center}
\includegraphics[width=1.0\linewidth]{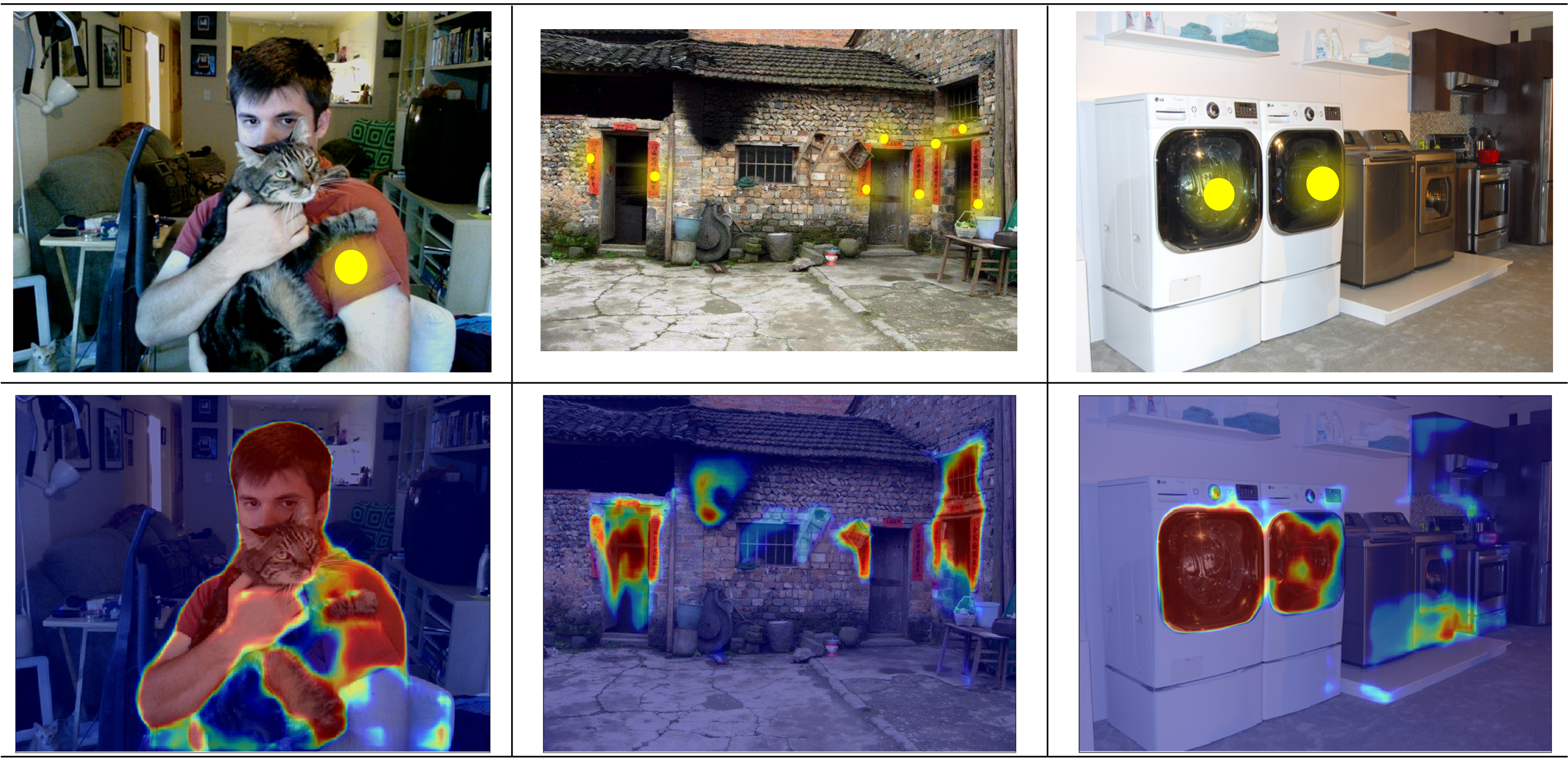}
\end{center}
   \caption{Visually salient objects do not always form the major events in the images and are not always what humans want to convey at first from the images. The yellow points in each image denote some visually salient objects. The saliency maps in the second row are obtained from \cite{HouCvpr2017Dss}. }
\label{fig:mot}
\end{figure}

\begin{figure}[htb]
\begin{center}
\includegraphics[width=1.0\linewidth]{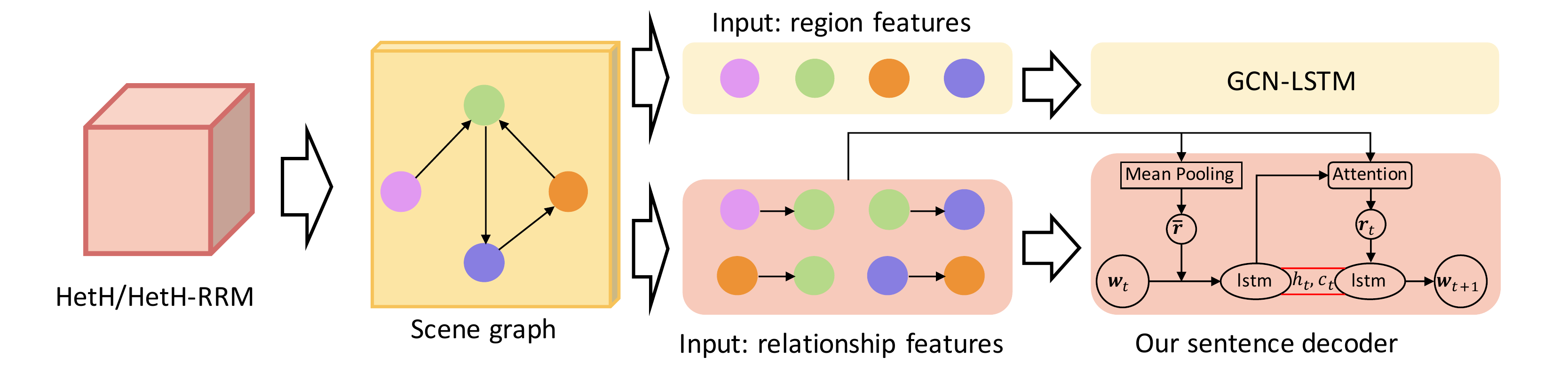}
\end{center}
   \caption{Our implementation of GCN-LSTM, and the the implementation scheme of our sentence decoder.}
\label{fig:dec}
\end{figure}

\section{Implemention Details for Image Captioning}\label{sec:captioning}
As the source codes of GCN-LSTM \cite{yao2018exploring} have not been released by the submission deadline, we re-implement it. In its original version, a simple two-layer MLP classifier is applied to predict the pairwise relationship, which acts as the frontend scene graph detector. For a fair comparison, we replace this detector with our HetH. To transform our HetH/HetH-RRM for image captioning task, we add a sentence decoder which is modified from LSTM backend of GCN-LSTM. The GCN-LSTM model conducts graph convolution on the scene graph and injects all relation-aware \textbf{region}-level features into a two-layer LSTM with attention mechanism. Different from GCN-LSTM, we intend to inject the relation features rather than region-level features, considering that the relationships which convey the events in the image are more helpful for description generation. In Figure \ref{fig:dec}, we show a brief diagram to illustrate our implementation of GCN-LSTM, and demonstrate the implementation scheme of our sentence decoder for image captioning.  

Specifically, we obtain a set of visual relationship representations $\{\bm{f}_m^\mathcal{R}\}_{m=1}^M$ ($\bm{f}_m^\mathcal{R}\in \mathbb{R}^D_f$, $D_f=4,096$) after relation context decoding (see \textcolor{blue}{Figure 2(d)} in the main paper). We concatenate them with the word embeddings of their subjects, objects, and predicates, denoted by $\bm{w}_m^s \in \mathbb{R}^{D_w}$, $\bm{w}_m^o \in \mathbb{R}^{D_w}$, and $\bm{w}_m^p \in \mathbb{R}^{D_w}$ ($D_w = 300$), and obtain $\{\bm{r}_m\}_{m=1}^M$ ($\bm{r}_m \in \mathbb{R}^{D_r}$, $D_r=4,996$):
\begin{equation}
\bm{r}_m = \left[\bm{f}_m^\mathcal{R}; \bm{w}_m^s; \bm{w}_m^o; \bm{w}_m^p\right].
\end{equation}

The sentence decoder is a two-layer LSTM. It's noted that two layers in this decoder share one hidden state $\bm{h}\in \mathbb{R}^{D_h}$ and cell state $\bm{c} \in \mathbb{R}^{D_h}$. At each time step $t$, the first layer collects the maximum contextual information by concatenating the input word embedding $\bm{w}_t \in \mathbb{R}^{D_w}$ and the mean-pooled visual relationship feature $\bar{\bm{r}}=\frac{1}{M}\sum\limits_{m=1}^M \bm{r}_m$. The updating procedure is as
\begin{equation}
\bm{h}_t^1, \bm{c}_t^1 = {f_1\left(\left[\bm{w}_t; \bar{\bm{r}}\right]\right)}_{|\bm{h}_{t-1}^2, \bm{c}_{t-1}^2},
\end{equation}where $f_1$ is the updating function within the first-layer unit, $|\bm{h}_{t-1}^2, \bm{c}_{t-1}^2$ denotes that the internal hidden state and cell state is the ones that updated by the second-layer unit from the previous timestep. Then we compute a normalized attention distribution over all the relationship features
\begin{equation}
a_{t, m} = \bm{W}_a\left[\tanh \left(\bm{W}_f\bm{r}_m + \bm{W}_h\bm{h}_t^1\right)\right], \,\, \lambda_t=\mathrm{softmax}(\bm{a}_t),
\end{equation}where $a_{t, m}$ is the $m$-th element of $\bm{a}_t$, $\bm{W}_a \in \mathbb{R}^{1 \times D_a}$, $\bm{W}_f \in \mathbb{R}^{D_a \times D_r}$, $\bm{W}_h \in \mathbb{R}^{D_a \times D_h}$ are transformation matrices. Specifically, both the dimension of the hidden layer $D_a$ for measuring the attention distribution and the dimension of the hidden layer $D_h$ in LSTM are set as 512.  $\lambda_t \in \mathbb{R}^M$ denotes the normalized attention distribution whose $m$-th element $\lambda_{t, m}$ is the attention weight of $\bm{r}_m$. The attended relationship feature is computed as $\bm{r}_t=\sum\limits_{m=1}^M \lambda_{t, m} \bm{r}_m$. Then the updating procedure of the second-layer unit is
\begin{equation}
\bm{h}_t^2, \bm{c}_t^2 = {f_2\left(\bm{r}_t\right)}_{|\bm{h}_{t}^1, \bm{c}_{t}^1},
\end{equation}where $f_2$ is the updating function within the second-layer unit. $\bm{h}_t^2$ is used to predict the next word through a softmax layer. 

\begin{table}
\caption{Statistics of VG200, VG-KR, and VG150.}
\begin{center}
\scalebox{0.7}{
\begin{tabular}{@{}lcccccccc@{}}\toprule
  Dataset & Images & \makecell[c]{Images with \\Relations} & \makecell[c]{Object\\ Categories} & \makecell[c]{Predicate \\ Categories} & \makecell[c]{Object\\ Instances} &  \makecell[c]{Relation\\ Instances} & \makecell[c]{Key Relation\\ Instances} & \makecell[c]{Images with Key \\ Relation Instances} \\\hline
  
  VG200 & 51,498 & 46,562 & 200 & 80 & 619,119 & 442,425 & \multirow{2}*{101,312} & \multirow{2}*{26,992} \\
  VG-KR & 26,992 & 26,992 & 200 & 80 & 360,306 & 250,755 \\\hline
  VG150 & 108,073 & 89,169 & 150 & 50 & 1,145,398 & 622,705 & - & - \\

\bottomrule
\end{tabular}}
\end{center}
\label{tab:datasets}
\end{table}

\begin{figure}[t]
\begin{center}
\includegraphics[width=1.0\linewidth]{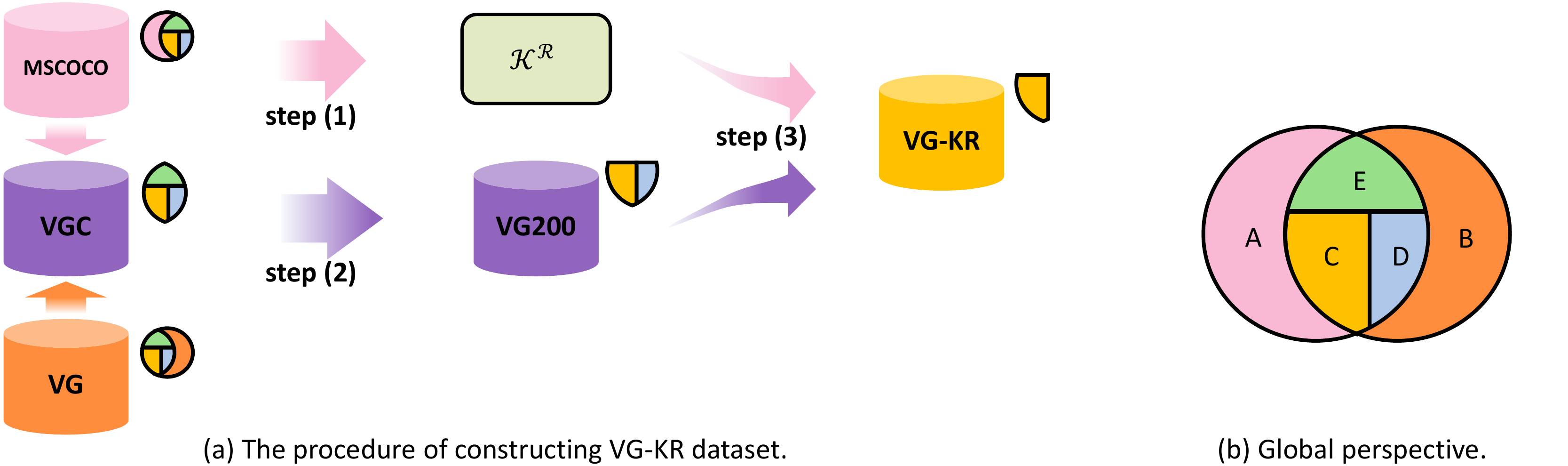}
\end{center}
   \caption{The procedure of VG-KR dataset construction. The color block shown in the top right of each dataset in (a) demonstrates its components. (b) gives a global perspective of these color blocks. Images in MSCOCO consist of four parts, A, C, D, and E. Images in VG consist of B, C, D, and E. Images in VGC consist of C, D, and E. E denotes images filtered in step (2) which do not contain any relation. D denotes images filtered in step (3) which do not contain any key relation. }
\label{fig:vg-kr-const}
\end{figure}

\begin{figure}[t]
\setlength{\abovecaptionskip}{-0.2cm}
\setlength{\belowcaptionskip}{-0.3cm}
\begin{center}
\includegraphics[width=1.0\linewidth]{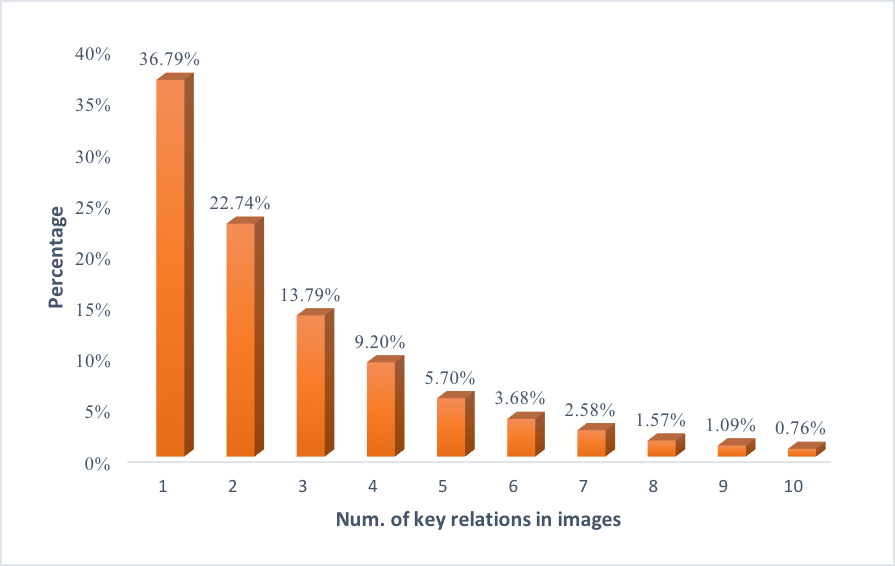}
\end{center}
   \caption{Distribution of images that contain different numbers of key relations.}
\label{fig:krdist}
\end{figure}

\begin{figure}[t]
\setlength{\abovecaptionskip}{-0.2cm}
\begin{center}
\includegraphics[width=1.0\linewidth]{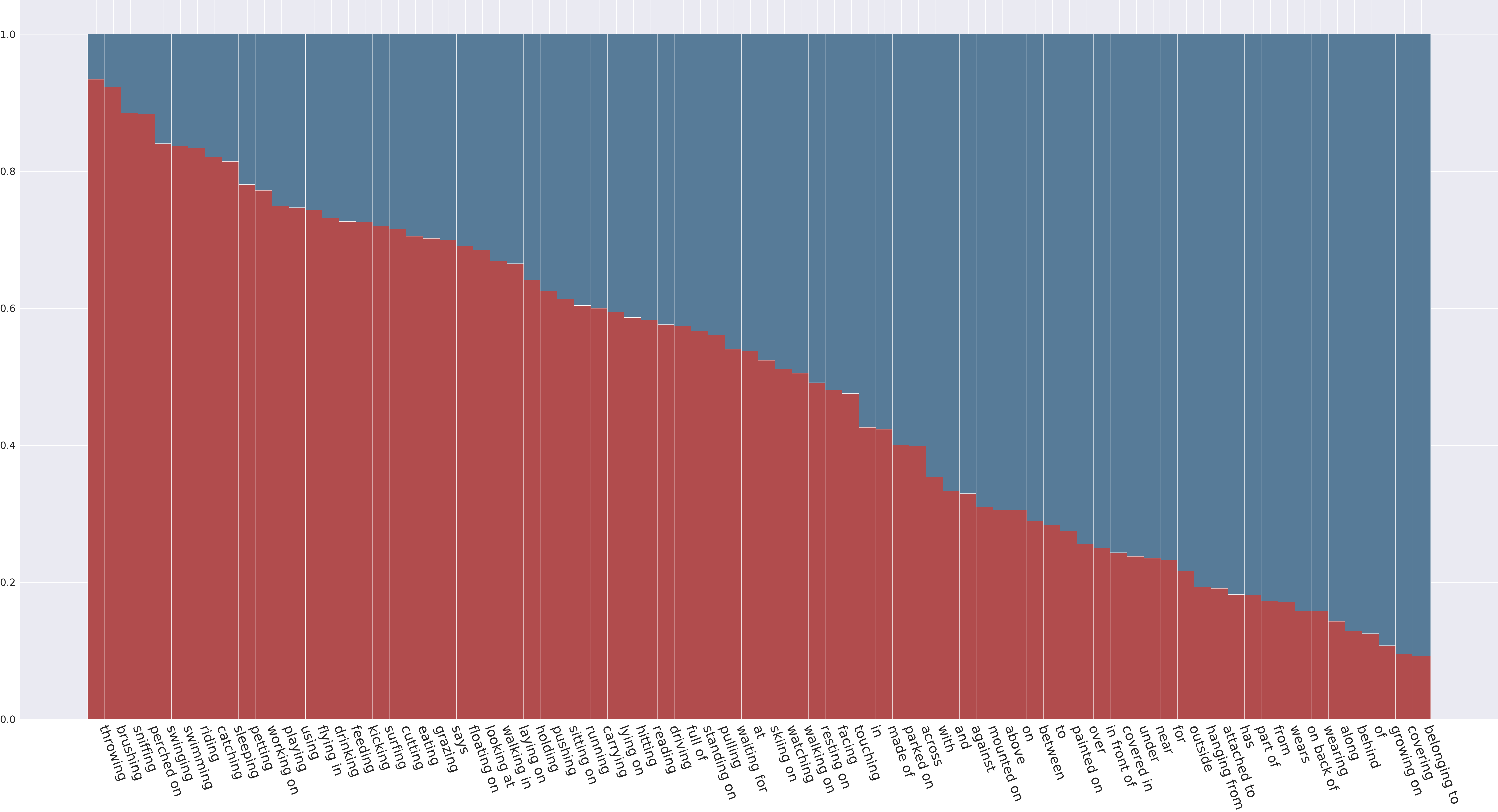}
\end{center}
   \caption{Distribution of the roles of a given predicate. The red bars stand for the probability of being key relations while the blue bars denote the probability of being the secondary ones. }
\label{fig:keyornon}
\end{figure}

\section{VG-KR Dataset Construction, Statistics, and Experimental Implementation Details}\label{sec:vgkr}

\subsection{VG-KR Dataset}

We demonstrate the procedure of constructing VG-KR dataset and different image sets involved in this procedure in Figure \ref{fig:vg-kr-const}. Concretely, 51,498 images in VG come from MSCOCO, which form the image set \textit{VGC}. We conduct three-stage processing on \textit{VGC}. (1) Stanford Scene Graph Parser \cite{schuster2015generating} is used to extract relation triplets from captions. They make up the set of key relations, denoted by $\mathcal{R^K}$. (2) We next cleanse the raw annotations of \textit{VGC} similar to \cite{xu2017scene}, keep the most frequent 150 object categories and 50 predicates, and add another most frequent 50 object categories and 30 predicates in $\mathcal{R^K}$, in order to keep as many key relations as possible for the following third step. After dropping images without relations in \textit{VGC}, we get a new subset \textbf{VG200} (\ie\,\,200 object categories) which contains 46,562 images. (3) Finally, we associate $\mathcal{R^K}$ with relation triplets in VG200 by associating their subject and object WordNet synsets \cite{Miller1992WORDNETAL} respectively. After filtering out the images without key relations in VG200, here comes \textbf{VG-KR} which contains 26,992 images. For both VG200 and VG-KR, we split the training and test set by 7:3 ratio, leading to 32,510/14,052 training/test images in VG200, and 18,720/8,272 training/test images in VG-KR. 

We show more detailed statistics and compare with VG150 in Table \ref{tab:datasets}. We can see that VG200 and VG-KR have more categories, as well as object and relation instances per image compared to VG150. Moreover, VG-KR contains indicative annotations of key relations. 

In Figure \ref{fig:krdist}, we show the distribution of images that contain different numbers of key relations. More than 90\% of images contain less than 5 key relations. It's reasonable because the key relations are obtained by matching the annotated relations with those extracted from captions. The number of relation triplets in captions generally is not very large. After all, a good caption is only requested to describe the major contents instead of the less important details. 

Given each predicate, we explore the distribution of its role, \ie, whether it belongs to a key relation or not. The result is shown in Figure \ref{fig:keyornon}. The predicates with large probability to be key ones, such as \textit{throwing}, \textit{brushing}, and \textit{sniffing}, are usually verbs containing rich semantics. They are image-specific and when we see these predicates, a scene can be roughly imagined. While predicates like \textit{belonging to}, \textit{of}, and \textit{behind}, which carry little information, are less likely to make up the key relations. 

\subsection{Settings and Implementation Details}

The dimension of hidden states and cells in both Hybrid-LSTM and RRM is 512. The sizes of $\bm{W}^{(r)}_1$ and $\bm{W}^{(r)}_2$ in \textcolor{blue}{Eq.(11)} in the main paper are $256\times 512$ and $1 \times 256$ respectively. The GloVe embedding vectors we use are of 200 dimensions. 

When training on the VG dataset, we follow previous works \cite{zellers2018neural,Tang_2019_CVPR} to extract the first 5,000 images of the training split and treat them as the validation split. The results reported on VG150, VG200, and VG-KR are obtained by firstly selecting the best model on validation split and then evaluating it on test split. As for the experiments on VRD, we report the results of the last epoch evaluated on test split without model selection (The hyperparameters settings are the same as those of experiments on VG).

We pre-train object detectors on VRD, VG150, and VG200 respectively and freeze the learned parameters. To train the whole model end-to-end, we use an SGD optimizer with a learning rate of 0.001 and the batch size is 10. When computing the ranking loss for RRM, we randomly sample 512 pairs of key triplets and secondary triplets. The margin $\gamma$ is empirically set to 0.5. All the existing methods evaluated on our VG200 or VG-KR datasets are retrained. 

\begin{figure}[t]
\begin{center}
\includegraphics[width=1.0\linewidth]{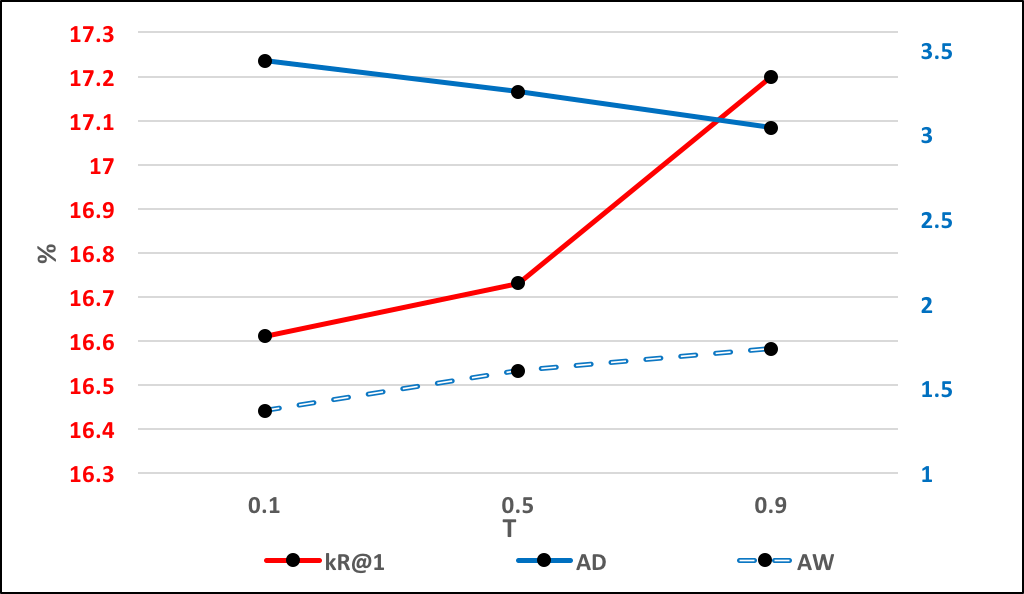}
\end{center}
   \caption{Effect of threshold $T$ when constructing HET. AD and AW denote average depth and average width respectively. The red curve stands for the kR@1 performance of HET-RRM, evaluated under P{\small{RED}}C{\small{LS}} protocol with triplet-match rule.}
\label{fig:thresh}
\end{figure}

The threshold $T$ for determining a parent node actually has direct influence on the shape of HET. We investigate the performance curve together with the  tree depth and width variation trend. As shown in Figure \ref{fig:thresh}, as $T$ varies from 0.1 to 0.9, the \lq\lq tall thin\rq\rq\,\,tree becomes a \lq\lq short fat\rq\rq\,\,tree, and the performance is improved. Thus we set $T$ to 0.9. 

As $T$ becomes larger, the condition that a node can be a parent node, \ie, $P_{nm} > T$ (\textcolor{blue}{Eq.(1)} in the main paper), is more and more difficult to be satisfied. Thus, our algorithm for constructing HET tends to set the root as the parent of a node, which results in a \lq\lq shorter\rq\rq\,\, and \lq\lq fatter\rq\rq\,\, tree. 

A small $T$ would lead to considerable wrong hierarchical connections. It's noted that the hierarchical connections in our HET have much stronger semantics than the associations of siblings. Therefore, a large $T$ eliminates wrong hierarchical connections as far as possible. Although it means that more entities are set as the child of the root and inappropriate siblings associations increase, proper hierarchical connections still plays a positive role in context encoding. 

\begin{table}
\caption{The results of multiple runs of HetH and the statistical significance. These results are obtained under the PREDCLS protocol. }
\begin{center}
\setlength{\tabcolsep}{8mm}{
\begin{tabular}{@{}cccc@{}}\toprule
  \#RUN & R@20 & R@50 & R@100\\\hline
  1 & 33.46 & 36.59 & 37.00 \\
  2 & 33.53 & 36.64 & 37.04 \\
  3 & 33.93 & 36.65 & 37.07 \\\hline
  $\mu\pm\sigma$ & 33.64$\pm$0.21 & 36.63$\pm$0.03 & 37.04$\pm$0.03\\

\bottomrule
\end{tabular}}
\end{center}
\label{tab:robust}
\end{table}

\section{Robustness Analyses}\label{sec:robust}
We make multiple runs on the HetH under the PREDCLS protocol. The results and statistical significance are shown in Table \ref{tab:robust}.

\begin{figure}[t]
\begin{center}
\includegraphics[width=1.0\linewidth]{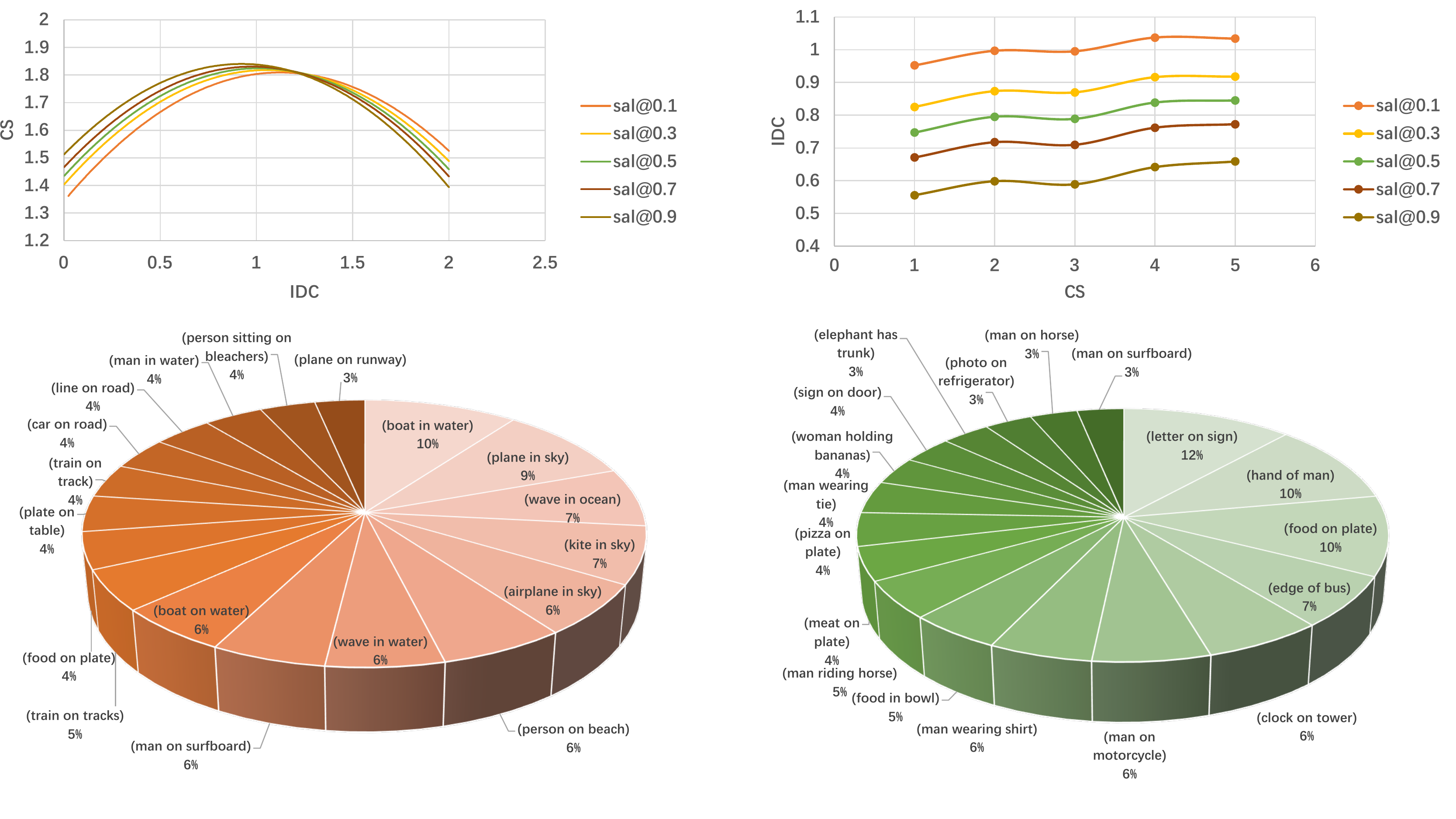}
\end{center}
   \caption{Curve charts and pie charts for the indicator which is the sum of subject and object visual saliency values. In the curve charts, different curves are drawn under different thresholds $T_s$. \textbf{Top left:} CS - IDC chart. \textbf{Top right:} IDC - CS chart. \textbf{Bottom left:} Component analysis for relations which have small IDC values. \textbf{Bottom right:} Component analysis for relations which have large IDC values.}
\label{fig:salchart}
\end{figure}

\begin{figure}[t]
\begin{center}
\includegraphics[width=1.0\linewidth]{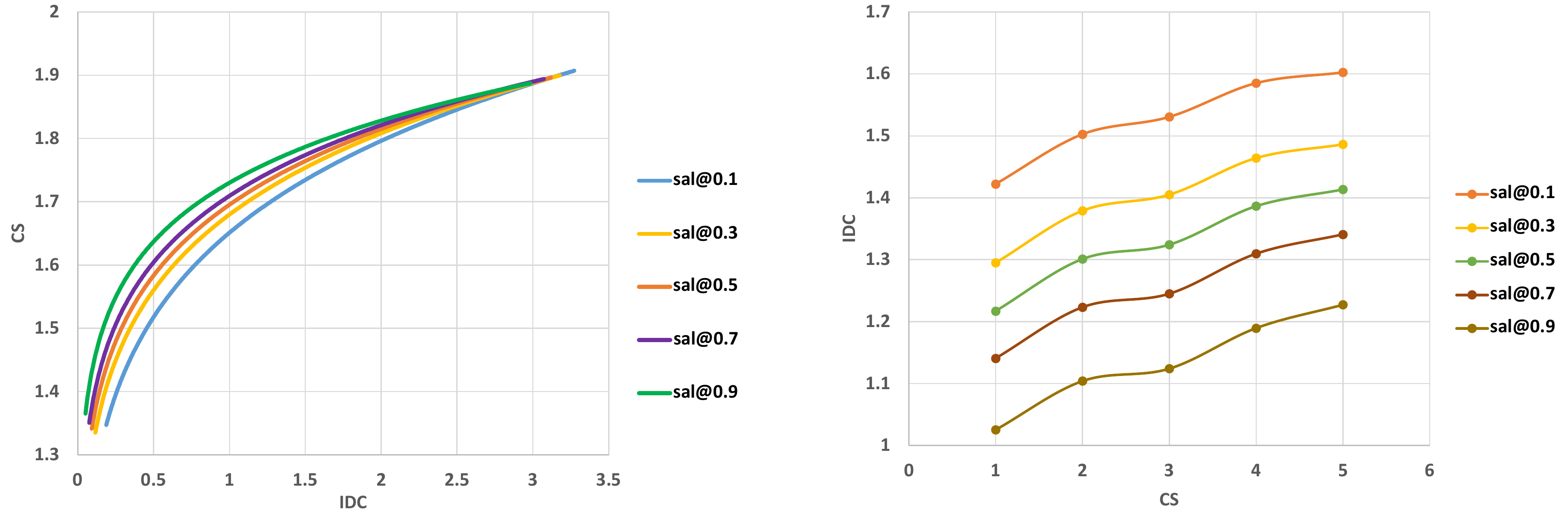}
\end{center}
   \caption{
   Curve charts for the indicator which is the sum of subject and object visual saliency values and normalized areas. Different curves are drawn under different thresholds $T_s$. \textbf{Left:} CS - IDC chart. \textbf{Right:} IDC - CS chart.}
\label{fig:salareachart}
\end{figure}

\section{Exploration on VG-KR}\label{sec:explor-vgkr}
We develop the Relation Ranking Module (RRM) to prioritize key relations. We intend to capture humans' subjective assessment on the importance of relations with some objective indicators. As analyzed in Section \ref{sec:mot}, visually salient objects engage humans' gaze and have the potential to form major events. Therefore, visual saliency can be one of the useful indicators. However, it's easy to lead to misunderstandings when only visual saliency is considered.

To better describe the importance of relation, we borrow the traditional \lq\lq saliency\rq\rq\, concept, and put forward a brand-new concept, \textit{cognitive saliency}, which tries to estimate the importance of a relation from humans' perspective as the sensation of importance of relation is very subjective. Considering the measurement of cognitive saliency of a relation triplet, we employ its \textit{times being referred within the five captions} of each image, which can be directly obtained during the construction of our VG-KR dataset. However, this measurement of cognitive saliency is not computable. (\ie, it is grading from humans, but cannot be directly used in computational models.) If we want to make use of the cognitive saliency, we need to find a computable indicator for it. The indicator should be proportional to cognitive saliency, which means that as the cognitive saliency goes up, the same trend should be observed on the indicator, and vice versa. 

Intuitively, the first possible indicator is the visual saliency of subject and object in a relation triplet. Specifically, we set the indicator $\Phi$ as the sum of saliency values of subject $o^{sub}$ and object $o^{obj}$:
 \begin{align}
\mathcal{S}^{sub}&=\frac{|\{p|p\in \bm{b}^{sub} \wedge \mathcal{S}^{p} > T_s\}|}{|\{p|p\in \bm{b}^{sub}\}|}, \\
\mathcal{S}^{obj}&=\frac{|\{p|p\in \bm{b}^{obj} \wedge \mathcal{S}^{p} > T_s\}|}{|\{p|p\in \bm{b}^{obj}\}|}, \\
\Phi&=\mathcal{S}^{sub} + \mathcal{S}^{obj}\label{eq:idc1},
 \end{align}
 where $p$ denotes pixels, $\bm{b}^{sub}$ and $\bm{b}^{obj}$ are bounding boxes of subject and object, $\mathcal{S}^p$ is the saliency value of pixel $p$, $\mathcal{S}^{sub}$ and $\mathcal{S}^{obj}$ are saliency values of subject and object, $T_s$ is a given threshold. $|\cdot|$ computes the number of elements in a set. The pixel-wise saliency is computed by one of the state-of-the-art saliency detectors \cite{HouCvpr2017Dss}. 

 To draw the Cognitive Saliency (CS, Y-axis) - Indicator (IDC, X-axis) curve chart, we randomly sampled 50,000 key relations from VG-KR with grading from 1 to 5 as their CS values. As IDC values (\ie, $\Phi$ in Eq. (\ref{eq:idc1})) are continuous, we sort all the sampled IDC values in ascending order, and divide them into 50 intervals $[\delta_k, \delta_{k+1}], 0\leq k \leq 50$, where $\delta_0=\rm{IDC}_{min}$ and $\delta_{50}=\rm{IDC}_{max}$. In each interval, we draw a point with the mean of the sampled IDC values as X-axis coordinate and the mean of sampled CS values as Y-axis coordinate. When it comes to IDC (Y-axis) - CS (X-axis) curve chart, the sampled relations are grouped by CS values. We compute the mean of IDC values for each group as Y-axis coordinates. These two charts are shown in Figure \ref{fig:salchart}. In each chart, we draw curves under different settings of $T_s$, denoted by $\mathrm{sal@}T_s$. From the IDC - CS chart at the top right of Figure \ref{fig:salchart}, IDC is proportional to CS. However, the CS - IDC chart at the top left of Figure \ref{fig:salchart} shows that CS is not strictly proportional to IDC, which means that although the computed visual saliency of an object is large, the relations involved in this object are not so important. What results in this phenomenon? We further extract the relations with relatively small IDC values and large IDC values respectively and analyze the ratio of each type of triplet. Concretely, we find the quartering points $\lambda_1 < \lambda_2 < \lambda_3$ of IDC values, and all the triplets whose IDC values are smaller than $\lambda_1$ or larger than $\lambda_3$ are picked out, namely the set $\Psi$ and $\Omega$. The component analysis results of the set $\Psi$ and $\Omega$ are shown at the bottom of Figure \ref{fig:salchart}, where the most frequent 18 types of triplets are demonstrated. 
 From the bottom left pie chart, lots of triplets with low IDC and low CS values generally are relations between relatively small objects and the large background entities. However, there are some exceptions, \eg, $\langle$\textit{man}, \textit{on}, \textit{surfboard}$\rangle$, and $\langle$\textit{train}, \textit{on}, \textit{tracks}$\rangle$. It's reasonable and we should explore the detailed image contents if we want to further analyze the association between their IDC and CS values. 
 What we should pay attention to is the bottom right pie chart, where we observe that most triplets in $\Omega$ are relations between an independent object and its components, such as $\langle$\textit{hand}, \textit{of}, \textit{man}$\rangle$ and $\langle$\textit{edge}, \textit{of}, \textit{bus}$\rangle$. Actually, these relations are indeed not so image-specific and carry little information. Humans generally overlook them. However, if the saliency of an object is large, saliency of its component will be large, too. It explains the phenomenon when IDC keeps increasing, the CS decreases instead. 

In order to further rectify the indicator above, we consider the size of subject and object out of the thinking that the sizes of components or details of a certain entity is relatively small, which can balance the large saliency value. Therefore, we add the normalized size of subject and object into the indicator:
\begin{equation}
\Phi' = \mathcal{S}^{sub} + \mathcal{S}^{obj} + \frac{A(o^{sub})}{A(o_{\mathcal{I}})} + \frac{A(o^{obj})}{A(o_{\mathcal{I}})},
\end{equation}
where $A(\cdot)$ denotes the size function, and $o_{\mathcal{I}}$ denotes the whole image. 
Similarly, we draw the IDC - CS and CS - IDC charts in Figure \ref{fig:salareachart}. It is shown that this improved indicator is a feasible one, as the CS is strictly proportional to IDC, and vice versa. 

The exploration above inspires us that an indicator which contains both the visual saliency and size of an object may be useful for finding key relations. Therefore, our devised RRM learns to capture humans' subjective assessment on the importance of relations under the guidance of visual saliency and entity size information.

\section{Additional Qualitative Results}\label{sec:results}
We demonstrate more qualitative results in Figure \ref{fig:overall_sg2}. From all of these examples, it can be seen that our RRM tends to describe relations between entities which are close to the root of HET. These relations describe the global contents and usually are what humans pay the most attention to. As a result, the captions generated from top relations better cover the essential contents. For example, in Figure \ref{fig:overall_sg2}(a), as the top-2 relations from HetH model contain $\langle$\textit{woman}, \textit{wearing}, \textit{boot\_1}/\textit{boot\_2}$\rangle$, the generated caption cannot capture the essential content that the \textit{woman} is \textit{holding} an \textit{umbrella}. On the contrary, top-2 relations from HetH-RRM successfully capture this information. In some cases, we observe that although top-2 relations do not contain the essential content, the generated caption can still capture it, \eg, the caption from HetH in Figure \ref{fig:overall_sg2}(b). It is mainly because the region of \textit{man\_1} contains part of the region of \textit{motorcycle\_1}, which provides visual cues for inferring the content that a \textit{man} is \textit{riding} a \textit{motorcycle}.

\begin{figure}
\setlength{\abovecaptionskip}{-0.2cm}
\begin{center}
\subfloat[]{\includegraphics[width=0.9\linewidth]{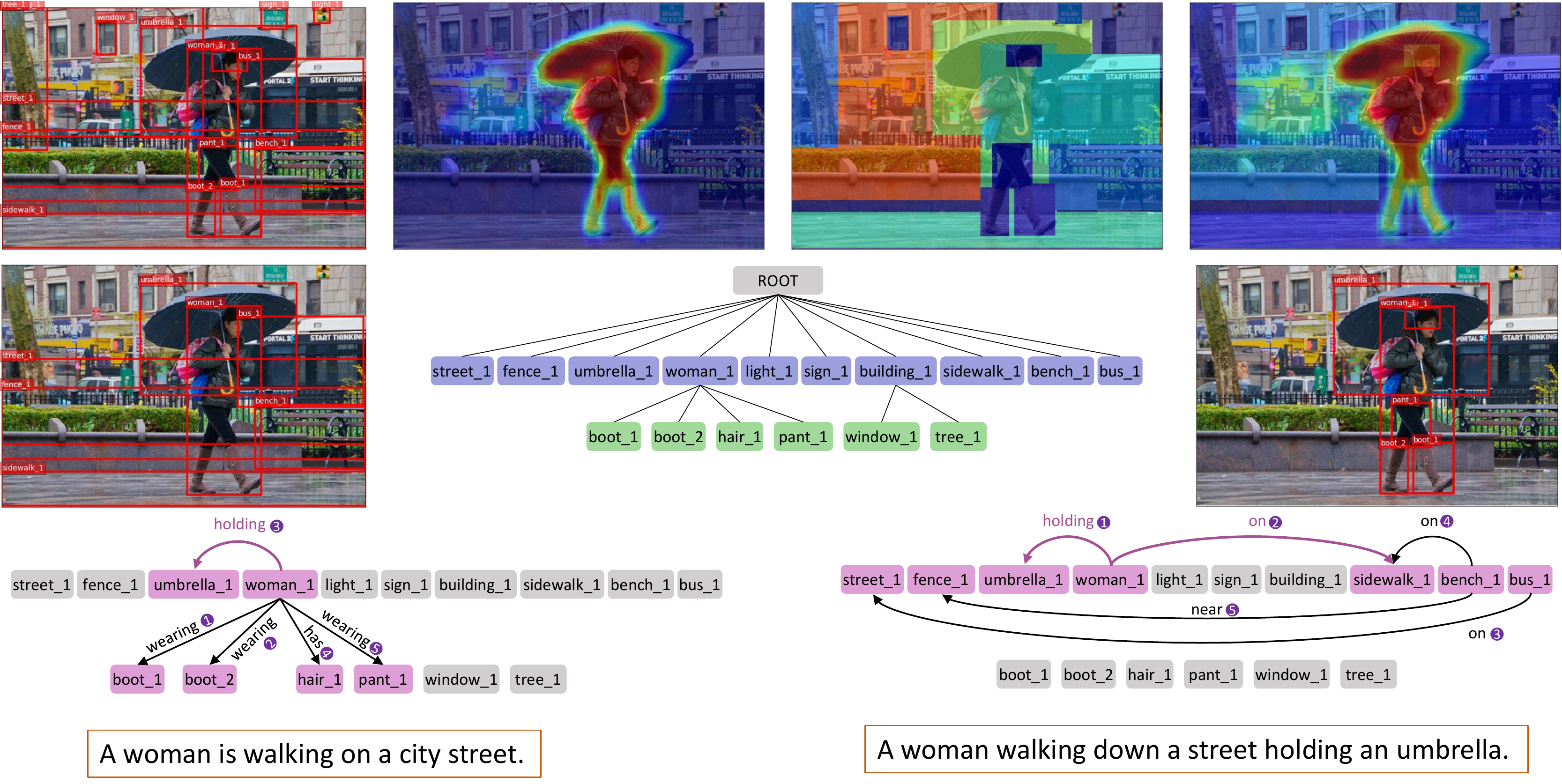}}\\
\subfloat[]{\includegraphics[width=0.9\linewidth]{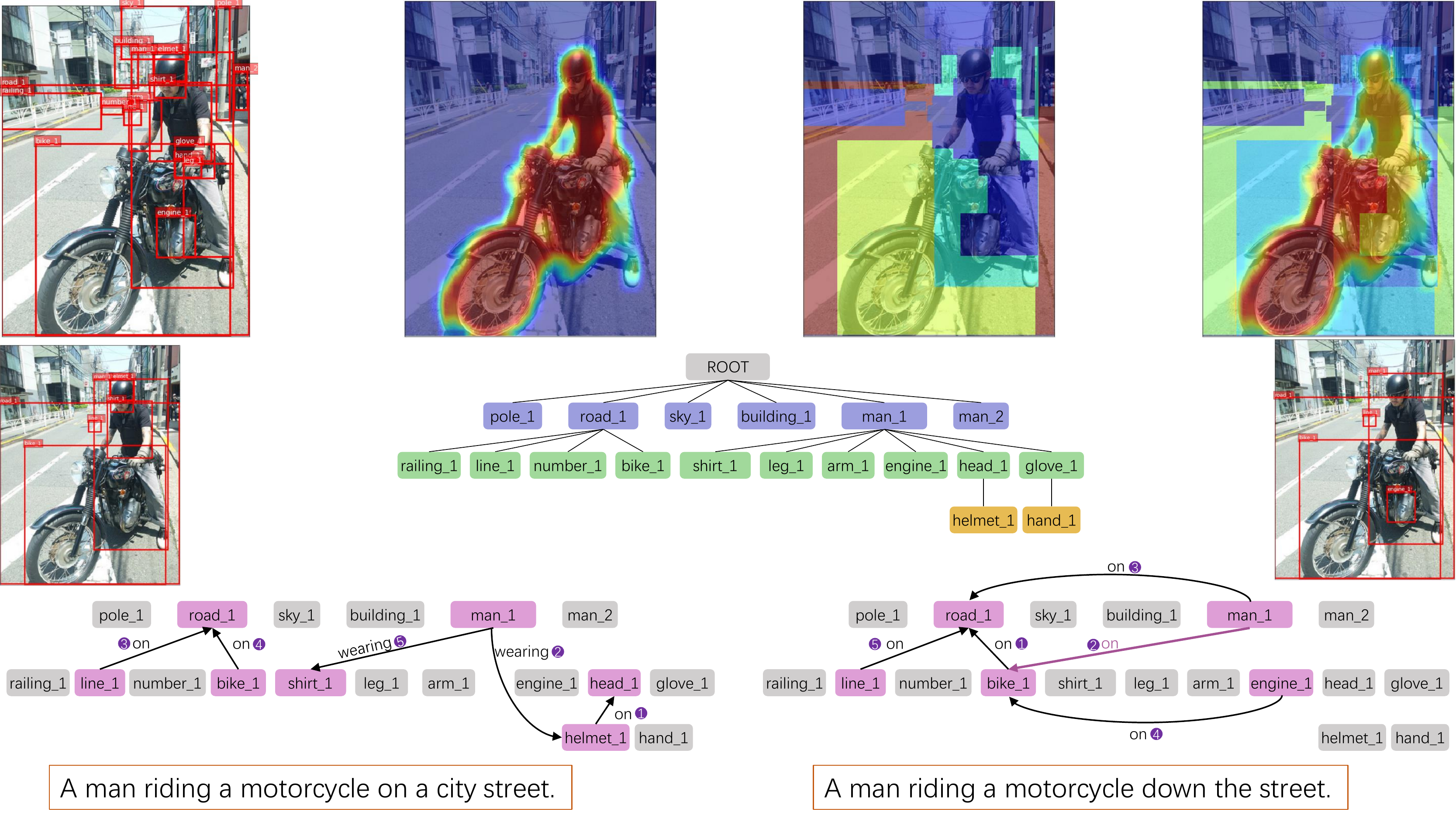}}\\
\end{center}
\end{figure}

\addtocounter{figure}{-1}
\begin{figure}
\addtocounter{figure}{1}
\addtocounter{subfigure}{2}
\setlength{\abovecaptionskip}{-0.2cm}
\begin{center}
\subfloat[]{\includegraphics[width=0.9\linewidth]{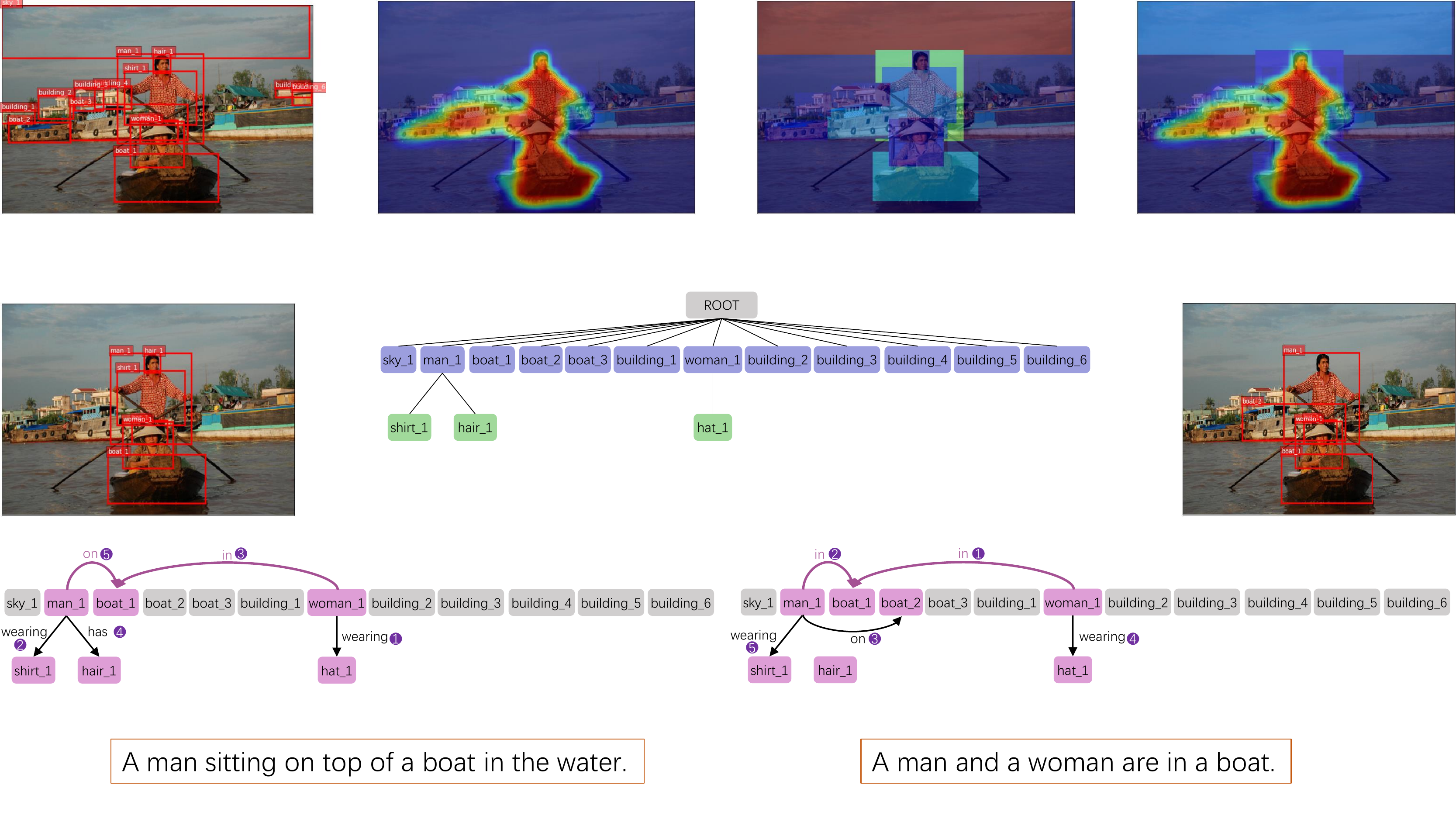}}\\
\subfloat[]{\includegraphics[width=0.9\linewidth]{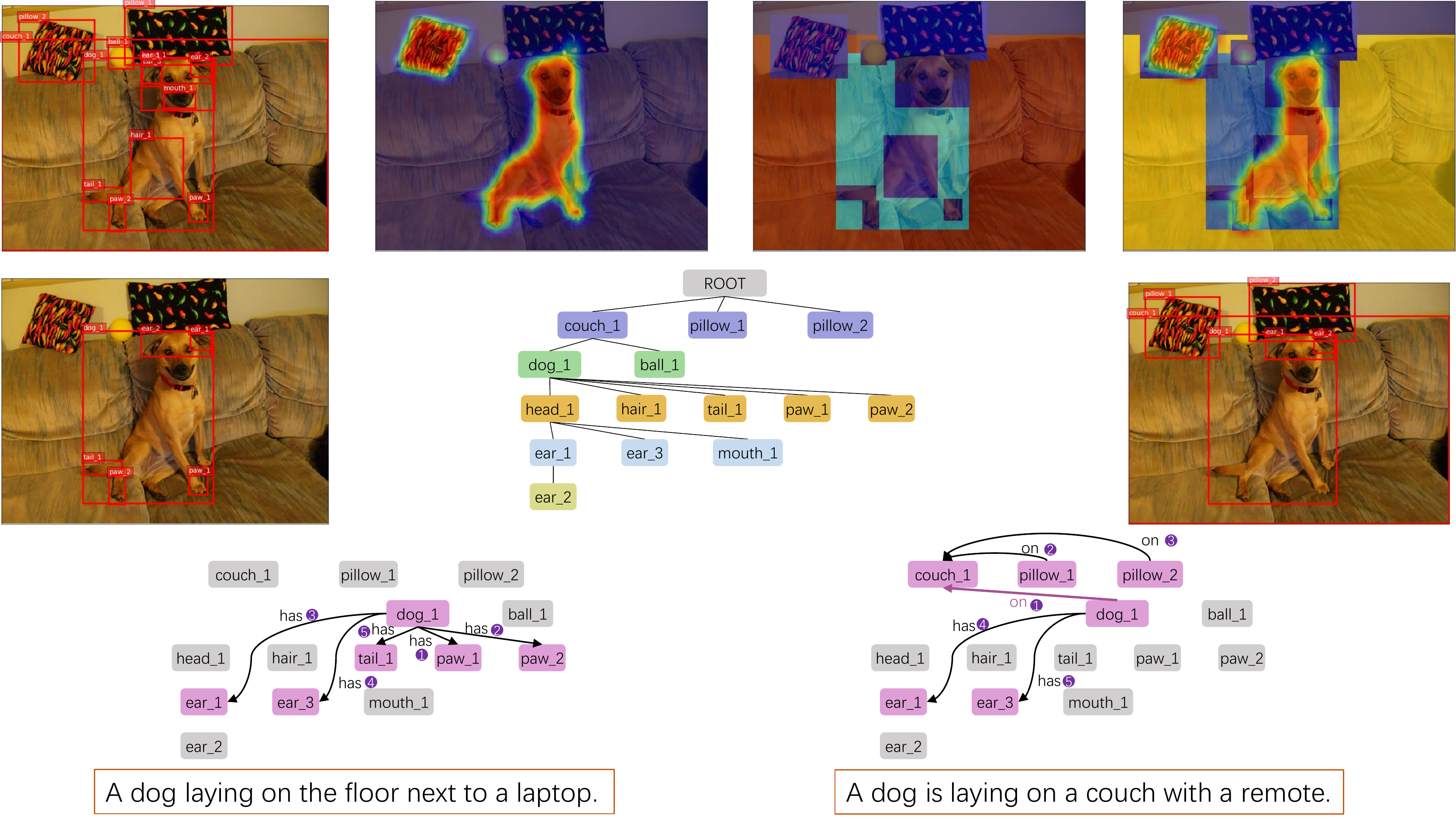}}\hspace{0pt}\\

\end{center}
   \caption{From top left to bottom right are: bounding boxes of all objects, saliency maps, area maps, mixed maps, bounding boxes of objects involved in top-5 relations from HetH, HET structure, bounding boxes of objects involved in top-5 relations from HetH-RRM model, hierarchical scene graphs from HetH and HetH-RRM model, generated captions using top-2 relations from HetH and HetH-RRM respectively. The purple arrows in scene graphs are key relations matched with ground truth. The purple numeric tags next to the relations are the rankings, and \lq\lq 1\rq\rq\,\, means that the relation gets the highest score.}
\label{fig:overall_sg2}
\end{figure}

\clearpage
%
%
\bibliographystyle{splncs04}
\bibliography{egbib}
\end{document}